%% file: main.tex
\documentclass{article}

\usepackage[preprint]{neurips_2026}

\usepackage[utf8]{inputenc} 
\usepackage[T1]{fontenc}    
\usepackage{hyperref}       
\usepackage{url}            
\usepackage{booktabs}       
\usepackage{amsfonts}       
\usepackage{nicefrac}       
\usepackage{microtype}      
\usepackage{xcolor}         

\usepackage{graphicx}
\usepackage{wrapfig}
\usepackage{amsmath}
\usepackage{adjustbox}
\usepackage{multirow}

\usepackage{subcaption}
\usepackage{colortbl}
\usepackage{longtable}
\usepackage{algorithm}

\usepackage{amssymb}
\usepackage{algorithmic}
\usepackage{mathtools}
\usepackage{amsthm}
\theoremstyle{plain}
\newtheorem{theorem}{Theorem}[section]
\newtheorem{proposition}[theorem]{Proposition}

\theoremstyle{definition}

\theoremstyle{remark}

\usepackage{thmtools}
\usepackage{thm-restate}

\title{MIDUS: Memory-Infused Depth Up-Scaling}

\author{%
  Taero Kim$^1$, Hoyoon Byun$^1$, Youngjun Choi$^1$, Sungrae Park$^2$, Kyungwoo Song$^1$\thanks{Corresponding Author}\\
  $^1$Yonsei University, $^2$Upstage AI\\
  \texttt{\{taero.kim,hoyun.byun,choiyj9803\}@yonsei.ac.kr,}\\
  \texttt{sungrae.park@upstage.ai, kyungwoo.song@yonsei.ac.kr}}

\begin{document}

\maketitle

\input{./sec/abstract}

\section{Introduction} \label{sec:intro}
\input{./sec/intro}

\section{Related Work} \label{sec:related}

\input{./sec/related}

\section{Preliminaries} \label{sec:preliminary}
\input{./sec/preliminaries}

\section{Methodology} \label{sec:method}
\input{./sec/methodology}

\section{Results} \label{sec:results}

\input{./sec/experiments}

\section{Conclusion} \label{sec:conclusion}
\input{./sec/conclusion}

{
\small
\bibliography{main}
\bibliographystyle{unsrt}
}


\appendix
\input{./sec/appendix}



\end{document}

%% file: sec/abstract.tex
\begin{abstract}
Expanding pre-trained language models offers a practical way to increase capacity without training larger models from scratch. Depth Up-Scaling (DUS) does so by duplicating Transformer blocks and inserting them into a pre-trained backbone. This process also duplicates FFN-heavy blocks, increasing parameter and compute cost while adding capacity through a block-level dense residual branch. Yet prior work suggests that added capacity need not remain tied to dense FFN branches, while attention heads often play heterogeneous roles, motivating more efficient head-level residual corrections. We propose Memory-Infused Depth Up-Scaling (MIDUS), which replaces the duplicated FFN branches with memory layers and turns added depth into lightweight retrieval-based residual capacity. We introduce a Head-wise Memory Layer (HML), which combines multi-head product-key memory with Head-wise Implicit Value Expansion (HIVE). HML assigns each head a distinct key space, while HIVE realizes head-specific values from a shared latent bank through compact projections. Alongside empirical improvements in performance and efficiency, our head-importance and fixed-retrieval structural analyses characterize HML with HIVE as a structurally distinct, head-conditioned alternative to FFN-based residual expansion.
\footnote[1]{Code: \url{https://github.com/MLAI-Yonsei/midus-hml}}
\end{abstract}

%% file: sec/intro.tex
\input{figure/fig1}

Large language models (LLMs) exhibit predictable performance gains as parameters and data scale, but training larger architectures from scratch remains prohibitively expensive. Model up-scaling therefore offers a practical alternative for reusing a pre-trained backbone by expanding it and adapting the resulting model through continual pre-training (CPT). Among up-scaling strategies, Depth Up-Scaling (DUS) increases computational depth while preserving the backbone structure by inserting additional Transformer blocks~\citep{gong2019efficient,yang2020progressively,wu2024llama}. Viewed through the residual architecture of Transformers, these inserted blocks act as new correction paths between pre-trained blocks. This makes their parameterization central to how additional capacity is introduced into the expanded model.

Recent work refines the placement and initialization of inserted blocks through selective block copying, learned initializers, or transport-guided initialization~\citep{wu2024llama,yang2025lesa,cao2025progressive}. However, these methods still duplicate FFN-heavy Transformer blocks, substantially increasing parameter and compute cost. The inserted FFN branch acts on the full hidden representation as a block-level dense update, tying added capacity to whole-block FFN replication. This motivates rethinking not only the cost of the inserted branch, but also the granularity at which its capacity is allocated within each inserted block.

More broadly, prior work suggests that Transformer capacity can be organized beyond fully dense FFN branches, either through conditional and structured computation or through sparse retrieval-based memory~\citep{fedus2022switch,zhang2022moefication,komatsuzaki2022sparse,lample2019large,kim2020large,huang2024ultra}. This perspective points to two directions for rethinking the inserted correction. First, memory layers offer efficient retrieval-based alternatives to FFNs, increasing model capacity with lower per-token or memory-access cost~\citep{lample2019large,kim2020large,huang2024ultra}. Second, attention heads differ in function and importance, with some heads contributing disproportionately and others specializing in distinct contextual patterns, suggesting that added capacity may benefit from head-dependent allocation~\citep{fernandez2024gradient,zhu2025focus,yin2025attention}. Together, these findings motivate a finer-grained view of DUS, where inserted capacity is both efficient and aligned with head-level heterogeneity.

In this paper, we propose Memory-Infused Depth Up-Scaling (MIDUS), a DUS framework that replaces inserted FFN branches with memory layers, turning added depth into lightweight retrieval-based capacity. To allocate this capacity within each inserted block, we introduce a Head-wise Memory Layer (HML), which consists of multi-head product-key memory and a value-side component, Head-wise Implicit Value Expansion (HIVE). HML uses attention-head outputs as memory queries and assigns each head a distinct product-key space, while HIVE realizes head-specific values from a shared latent value bank through compact projections. The resulting MIDUS-HML parameterizes the added residual branch as head-conditioned memory-based correction.

Empirically, MIDUS-HML improves both performance and efficiency over DUS baselines on Llama-3.2-1B and Llama-3.1-8B~\citep{grattafiori2024llama} under CPT and supervised fine-tuning (SFT). The gains hold across general-purpose, math-domain, retrieval-heavy, and long-context evaluations, while requiring fewer parameters, lower peak GPU memory, and lower computational cost. A head-importance study and a fixed-retrieval structural analysis further support MIDUS-HML as an effective head-conditioned alternative to FFN-based residual expansion.

Our contributions are threefold. First, we propose MIDUS, a DUS framework that replaces inserted FFN branches with memory layers, turning added depth into lightweight retrieval-based capacity. Second, we introduce HML, which combines head-wise product-key retrieval with HIVE for compact head-specific value realization. Third, we provide a structural analysis of the added residual branch, showing that, under a fixed-retrieval setup, head-specific value realization through HIVE recovers the head-specific optimum while a shared realized value bank incurs an irreducible gap.

%% file: figure/fig1.tex
\begin{wrapfigure}{r}{0.42\textwidth}
    \vspace{-1em}
    \centering
    \includegraphics[width=0.95\linewidth]{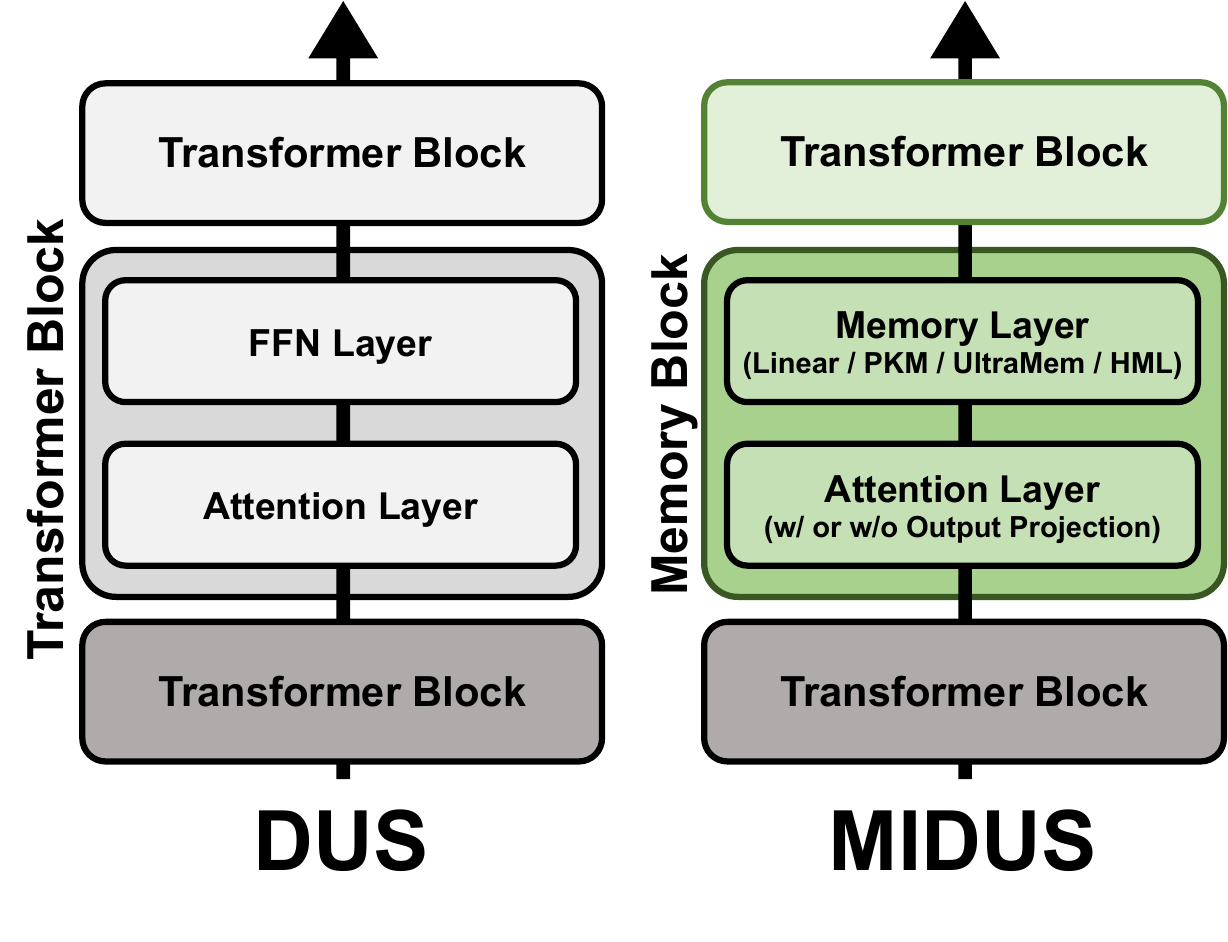}
    \vspace{-1.3em}
    \caption{DUS expands model depth through duplicating Transformer blocks, while MIDUS integrates Memory blocks.}
    \vspace{-1em}
    \label{fig:fig1}
\end{wrapfigure}

%% file: sec/related.tex
\subsection{Depth Up-Scaling}

Model up-scaling expands capacity along different axes, such as the width-wise route taken by MoE that selectively activates expert FFNs~\citep{shazeer2017outrageously,jiang2024mixtral}. DUS instead expands models depth-wise by duplicating and reorganizing Transformer blocks. Early work studies block selection and growth schedules through stacking and progressive training~\citep{gong2019efficient,yang2020progressively,du2024stacking,saunshi2024inductive,pan2024preparing,yano2025efficient}. Llama Pro selectively copies blocks under a stabilizing recipe~\citep{wu2024llama}, while SOLAR stacks contiguous block groups~\citep{kim2023solar}. Recent methods focus on initialization. LESA learns per-layer initializers~\citep{yang2025lesa}, while OpT-DeUS uses transport-guided initialization and Avg-DeUS initializes new layers by averaging adjacent blocks~\citep{cao2025progressive}.

Despite this progress, DUS still adds capacity through FFN-heavy Transformer blocks, increasing compute and memory with duplicated depth. Token-routing methods~\citep{raposo2024mixture,tan2024dlo} alleviate runtime cost but still keep FFN parameters at inference. MIDUS instead changes the inserted block itself, replacing duplicated FFN branches with memory layers, so the added depth carries lightweight retrieval-based capacity rather than additional dense FFN branches.

\subsection{Memory Layer} 
A memory layer stores key--value slots and retrieves values through dot-product lookup, forming a linear memory in its basic form. Linear memories scale poorly because lookup cost and key parameters grow linearly with memory size. PKM~\citep{lample2019large} factorizes the key space into two codebooks, enabling lookup over many composite keys with fewer dot products and improving compute and parameter efficiency over linear memory. Prior work uses PKM to augment or replace FFNs in pre-training, enabling retrieval-based computation in place of wide dense layers~\citep{kim2020large}. Ultra-sparse memory (UltraMem)~\citep{huang2024ultra,huang2025ultramemv2} further expands memory capacity through implicit value expansion and compressed representations while keeping per-token compute low.

MIDUS brings this memory-layer view to DUS by replacing the FFN branches that DUS would otherwise duplicate in inserted blocks. HML allocates memory capacity at the attention-head level by combining head-wise retrieval with HIVE-based value realization. Each head queries a distinct key space, while HIVE compactly realizes head-specific values from a shared latent bank. This aligns inserted memory capacity with head-level heterogeneity while keeping value storage tractable.

\subsection{Specialization and Importance of Attention Heads} 
Attention heads specialize and vary in importance across layers, inputs, and tasks. Prior analyses show that many heads can be pruned with limited degradation, while a smaller subset plays a disproportionate role in model behavior~\citep{michel2019sixteen,voita2019analyzing}. Other studies identify functional patterns such as local syntax, long-range dependency tracking, and task-dependent context use~\citep{clark2019does,fernandez2024gradient,yin2025attention}. Recent attribution analyses further show highly uneven head contributions, with a few heads dominating specific decisions while many remain inactive~\citep{zhu2025focus}. These findings suggest that added capacity need not be allocated only through block-level dense updates. HML instead allocates memory capacity at the attention-head level, giving each head a distinct key space and head-specific value realization.

%% file: sec/preliminaries.tex
We begin by formalizing the standard pre-norm Transformer block \citep{xiong2020layer} and a linear memory layer.

\paragraph{Transformer block.}
A pre-norm Transformer block is a mapping
$B \colon \mathbb{R}^{T \times d} \to \mathbb{R}^{T \times d}$. Given
$x \in \mathbb{R}^{T \times d}$, it is defined as
\begin{equation}
B(x) = a(x) + \mathrm{FFN}\big(\mathrm{LN}(a(x))\big), \quad \text{where} \ a(x) = x + \mathrm{Attn}\big(\mathrm{LN}(x)\big).
\end{equation}
$\mathrm{LN}$ denotes layer normalization and $\mathrm{Attn}$ the attention layer. The block applies an attention residual update followed by an FFN residual update. We denote the $\ell$-th Transformer block by $B_\ell$.

\paragraph{Linear memory layer.}
Given an input representation $a \in \mathbb{R}^{T \times d}$, a memory layer $\mathrm{Mem} \colon \mathbb{R}^{T \times d} \to \mathbb{R}^{T \times d}$ returns a retrieval-based representation $m = \mathrm{Mem}(a)$. 
It is specified by a query map and a key--value memory bank. For a \textit{linear} memory layer, define the query
$q(a)=aW_q\in\mathbb{R}^{T\times d_q}$ with learnable query projection $W_q\in\mathbb{R}^{d\times d_q}$, 
keys $K\in\mathbb{R}^{N\times d_q}$, and values $V\in\mathbb{R}^{N\times d}$ for $N$ memory slots.
The similarity score matrix $S(a) \in \mathbb{R}^{T \times N} $ is
\begin{equation}
S(a) = q(a)K^\top. \nonumber
\end{equation}
For each token $t \in \{1,\dots,T\}$, let $S(a)_t \in \mathbb{R}^{N}$ denote
the $t$-th row of $S(a)$. Let $\Omega_t \subset \{1,\dots,N\}$ be the Top-$k$
index set of $S(a)_t$, with $|\Omega_t|=k$. Let $S(a)_{t,\Omega_t}$ denote
$S(a)_t$ restricted to the indices in $\Omega_t$. We define
$\alpha_t \in \Delta^{N-1}$ as the softmax over $S(a)_{t,\Omega_t}$ with zero
mass outside $\Omega_t$. For the retrieved representation $m \in \mathbb{R}^{T\times d}$, each row is computed as
\[
m_t = \sum_{j=1}^{N} \alpha_{t,j} V_j,
\qquad t=1,\dots,T. \nonumber
\]

%% file: sec/methodology.tex
To parameterize the added residual capacity in DUS without duplicating FFN-heavy blocks, we introduce MIDUS and HML. In MIDUS, a Memory block replaces the FFN branch of an inserted Transformer block with a memory layer, so that the added residual correction is carried by retrieval-based capacity rather than by a duplicated feed-forward branch. We first describe the MIDUS interleaving rule, initialization, and residual path that preserve the base model at training start. We then introduce HML, a head-wise memory design built on product-key lookup and efficient value factorization. We focus on the architectural formulation below and defer implementation-level optimizations for efficient HML training and inference to Appendix~\ref{appendix:implement}.

\subsection{Memory-Infused Depth Up-Scaling}
\label{sec:midus_method}

\paragraph{Memory block.}
A Memory block $M : \mathbb{R}^{T \times d} \to \mathbb{R}^{T \times d}$ replaces the FFN branch in a Transformer block with the memory layer $\mathrm{Mem}$.
\begin{equation}
M(x) = x + \mathrm{Mem}(a(x)),
\label{eq:mem_block}
\end{equation}
where $a(x)$ is an attention-updated representation. The attention layer first contextualizes tokens across the sequence, and the memory layer then produces a retrieval-based residual correction by querying a memory bank and aggregating selected values. Substituting a small number of FFNs with memory layers has been shown to improve pre-training performance~\citep{lample2019large}. We build on this idea for DUS, where added depth is carried by retrieval-based capacity rather than duplicated FFN branches.

\paragraph{Memory-Infused Depth Up-Scaling.}
Let $f_0(x)=B_{L-1}\circ\cdots\circ B_0(x)$ be the base model. DUS
selects anchor indices $0\le i_0<\cdots<i_{D-1}\le L-1$ and interleaves
$D$ additional Transformer blocks
$\widetilde B_0,\ldots,\widetilde B_{D-1}$ after the corresponding anchor
blocks~\citep{wu2024llama}, yielding
\begin{equation}
f_{\mathrm{DUS}}(x)
=
B_{L-1}\circ\cdots\circ
\widetilde B_{D-1}\circ B_{i_{D-1}}
\circ\cdots\circ
\widetilde B_0\circ B_{i_0}
\circ\cdots\circ B_0(x).
\label{eq:dus}
\end{equation}
MIDUS uses the same anchor indices, but inserts Memory blocks
$M_0,\ldots,M_{D-1}$ before the corresponding anchor blocks:
\begin{equation}
f_{\mathrm{MIDUS}}(x)
=
B_{L-1}\circ\cdots\circ
B_{i_{D-1}}\circ M_{D-1}
\circ\cdots\circ
B_{i_0}\circ M_0
\circ\cdots\circ B_0(x).
\label{eq:midus}
\end{equation}

\input{figure/fig2}

The ordering difference between Eq.~\eqref{eq:dus} and Eq.~\eqref{eq:midus} is deliberate and distinguishes MIDUS from standard interleaved DUS~\citep{wu2024llama}. Placing each Memory block before the corresponding pre-trained block aligns it with the block used for attention initialization, as described below. In this work, we do not consider DUS variants that deepen models by stacking contiguous groups of Transformer blocks, as in \citep{kim2023solar}. Figure~\ref{fig:fig1} shows the structural differences between DUS and MIDUS. Appendix~\ref{appendix:policy} details the placement policies used by different methods and their implications for training efficiency.

\paragraph{Weight initialization and residual connection.}
Recall from Eq.~\eqref{eq:mem_block} that a Memory block outputs $M(x)=x+m$ with $m=\mathrm{Mem}(a)$ constructed from retrieved values. We initialize the value bank to zero so that $\mathrm{Mem}(a)=0$ at initialization and the block becomes the identity $M(x)=x$. This ensures that the expanded model matches the base model at the start of training and provides a stable starting point for the memory parameters. In addition, under the placement in Eq.~(4), each Memory block initializes its attention parameters from the corresponding pre-trained Transformer block that follows it. Since the Memory block is placed before this block, it forms an attention-updated representation aligned with the context processed by the following pre-trained block, rather than an unrelated transformation. This provides a stable basis for learning a retrieval-based residual update, which is incorporated through the residual path in Eq.~\eqref{eq:mem_block}.

\subsection{Head-wise Memory Layer}
\label{sec:hml}

HML parameterizes the memory-based residual correction at the granularity of attention heads. Motivated by evidence that attention heads specialize in distinct functions~\citep{fernandez2024gradient,zhu2025focus,yin2025attention}, HML generates head-specific queries and assigns each head a distinct key space and value realization. This aligns inserted capacity with head-level representations while retaining efficient sparse lookup. HML combines multi-head product-key memory for efficient head-wise key-space allocation with HIVE for compact head-specific value realization.

\paragraph{Multi-head product-key memory.}
Let $H$ denote the number of attention heads. HML assigns one memory head to
each attention head, so that each attention-head output queries a distinct key
space. This makes retrieval head-aligned rather than shared across heads. A naive design with an independent linear key bank per head would make head-wise retrieval costly, since both lookup cost and key parameters scale with the number of slots. We therefore use a PKM structure~\citep{lample2019large}, retaining its row--column key factorization and two-stage Top-$k$ selection within each memory head.

We first define the product-key lookup rule for a generic query representation \(q(a)\). This will later be specialized to HML by setting each query slice \(q_h\) to the attention-head output \(a_h(x)\). Consider \(q(a) \in \mathbb{R}^{T\times d_q}\) partitioned into \(H\) slices \(q_h\in\mathbb{R}^{T\times 2d_p}\) with \(d_p=d_q/(2H)\). We split each slice as \(q_h=[\,q_h^{\mathrm{row}}\mid q_h^{\mathrm{col}}\,]\), where \(q_h^{\mathrm{row}},q_h^{\mathrm{col}}\in\mathbb{R}^{T\times d_p}\). For each slice \(h\), we maintain sub-key banks \(K_h^{\mathrm{row}},K_h^{\mathrm{col}}\in\mathbb{R}^{n\times d_p}\) whose Cartesian product yields \(N=n^2\) composite keys indexed by \(\pi(i,j)=(i-1)n+j\). The row and column scores are
\begin{equation}
S_h^{\mathrm{row}}
=
q_h^{\mathrm{row}}{K_h^{\mathrm{row}}}^{\!\top},
\qquad
S_h^{\mathrm{col}}
=
q_h^{\mathrm{col}}{K_h^{\mathrm{col}}}^{\!\top},
\qquad
S_h^{\mathrm{row}},S_h^{\mathrm{col}}\in\mathbb{R}^{T\times n}.
\nonumber
\end{equation}

For each head $h$ and token $t$, let
$I_{h,t},J_{h,t}\subset\{1,\ldots,n\}$ be the Top-$k$ index sets of
$S_{h,t}^{\mathrm{row}}$ and $S_{h,t}^{\mathrm{col}}$, respectively, with
$|I_{h,t}|=|J_{h,t}|=k$. These sets form the $k^2$ candidate pairs
\begin{equation}
\widehat{\Omega}_{h,t}=I_{h,t}\times J_{h,t}.
\nonumber
\end{equation}
For each candidate pair $(i,j)\in\widehat{\Omega}_{h,t}$, we define the
additive pair score
\begin{equation}
\sigma_{h,t}(i,j)
=
S_{h,t}^{\mathrm{row}}(i)+S_{h,t}^{\mathrm{col}}(j).
\nonumber
\end{equation}
For each head $h$ and token $t$, let
$\Omega_{h,t}\subset\widehat{\Omega}_{h,t}$ be the final Top-$k$ pair set
under the pair scores $\sigma_{h,t}$, with $|\Omega_{h,t}|=k$. We define
$\alpha_{h,t}$ as the softmax weights over the selected pair scores
$\{\sigma_{h,t}(i,j)\}_{(i,j)\in\Omega_{h,t}}$. With the shared value bank 
$V\in\mathbb{R}^{N\times d}$, the retrieved output for head $h$ and token $t$
is
\begin{equation}
M_{h,t}(a)
=
\sum_{(i,j)\in\Omega_{h,t}}
\alpha_{h,t}(i,j)\,
V_{\pi(i,j)}
\in\mathbb{R}^{d}. \nonumber
\end{equation}
Stacking over tokens gives $M_h(a)\in\mathbb{R}^{T\times d}$, and aggregating
heads gives $m=\sum_{h=1}^{H}M_h(a)$.

Compared with a multi-head linear memory, multi-head PKM preserves $N$
addressable composite keys per head while reducing score-computation cost and
key parameters. A multi-head linear memory computes scores for all $T$ tokens
at cost $\mathcal{O}(TNd_q)$. PKM instead uses two $n$-way score computations
per head with slices of width $d_p=d_q/(2H)$, yielding
$\mathcal{O}(Tnd_q)$, a $\sqrt{N}$ reduction since $N=n^2$. Key parameters
drop from $Nd_q$ to $nd_q$, as each head stores only $2n$ sub-keys of width
$d_p$.

However, this formulation still uses a single realized value bank $V$ across heads,
so different heads retrieve from the same output value space despite distinct
roles. This can limit head-wise specialization in retrieved content. Inspired
by implicit-value designs~\citep{huang2024ultra}, we introduce Head-wise
Implicit Value Expansion (HIVE), which separates shared latent storage from
head-specific value realization.

\paragraph{Head-wise Implicit Value Expansion.}
Let $d_h=d/H$ and let $r$ denote the latent value dimension. Instead of sharing
a single realized value bank $V$ across heads, HIVE
introduces a shared \textit{latent} value table
$\bar V\in\mathbb{R}^{N\times r}$ and head-specific value projections
$W_h\in\mathbb{R}^{r\times d_h}$ for $h\in\{1,\dots,H\}$. Reusing the selected pairs $\Omega_{h,t}$ and
weights $\alpha_{h,t}$, HIVE first retrieves a latent value for token $t$ and
head $h$:
\begin{equation}
\bar M_{h,t}
=
\sum_{(i,j)\in\Omega_{h,t}}
\alpha_{h,t}(i,j)\,
\bar V_{\pi(i,j)}
\in\mathbb{R}^{r},
\qquad
M^{\mathrm{HIVE}}_{h,t}
=
\bar M_{h,t} W_h
\in\mathbb{R}^{d_h}.
\label{eq:hive}
\end{equation}
Stacking over tokens gives $M_h^{\mathrm{HIVE}}\in\mathbb{R}^{T\times d_h}$,
and concatenating heads gives
\begin{equation}
m=
\big[
M_{1}^{\mathrm{HIVE}}
\mid
M_{2}^{\mathrm{HIVE}}
\mid
\cdots
\mid
M_{H}^{\mathrm{HIVE}}
\big]
\in\mathbb{R}^{T\times d}.
\nonumber
\end{equation}
This factorization shares latent value storage while allowing head-specific value realization. A naive head-wise value bank requires $HNd_h$ parameters, whereas HIVE uses $Nr+Hrd_h$, reducing to $Nd_h+Hd_h^2$ when $r=d_h$ and becoming smaller when $N\gg d_h$. Its only added computation is a head-specific projection on the retrieved latent values, rather than a full-dimensional operation.

\paragraph{Head-wise Memory Layer.}
In HML, we use the pre-output-projection attention output as the memory input, preserving head-wise information before lookup. Let
\begin{align}
a'(x) &= \mathrm{Attn}'\!\big(\mathrm{LN}(x)\big)
     =
     \big[
     a_1(x)
     \mid
     a_2(x)
     \mid
     \cdots
     \mid
     a_H(x)
     \big],
     \nonumber
\end{align}
where $\mathrm{Attn}'$ denotes multi-head attention without the output projection and $a_h(x) \in \mathbb{R}^{T\times d_h}$ is the output of head $h$. Since each $a_h(x)$ already provides a head-specific query, HML sets $q_h := a_h(x)$ and omits the memory query projection $W_q$. This instantiates the product-key decomposition above with $d_q=d$ and $d_p=d_h/2$, so each head output is split into row and column query parts.

We denote by $\mathrm{HML}\colon \mathbb{R}^{T\times d}\to\mathbb{R}^{T\times d}$ the head-wise memory layer that combines the multi-head product-key memory with HIVE value projection above. The resulting Memory block is
\begin{equation}
M^{\mathrm{HML}}(x) = x + \mathrm{HML}\!\big(a'(x)\big). \nonumber
\end{equation}
To equip MIDUS with HML, we instantiate the inserted Memory blocks with $M^{\mathrm{HML}}$ in Eq.~\eqref{eq:midus}, yielding the MIDUS-HML configuration. Figure~\ref{fig:fig2} illustrates its step-by-step operation.

\subsection{Structural Analysis of Head-wise Value Parameterization}
\input{./sec/theory}

%% file: figure/fig2.tex
\begin{figure*}[t!]
    \centering
    \includegraphics[width=0.85\linewidth]{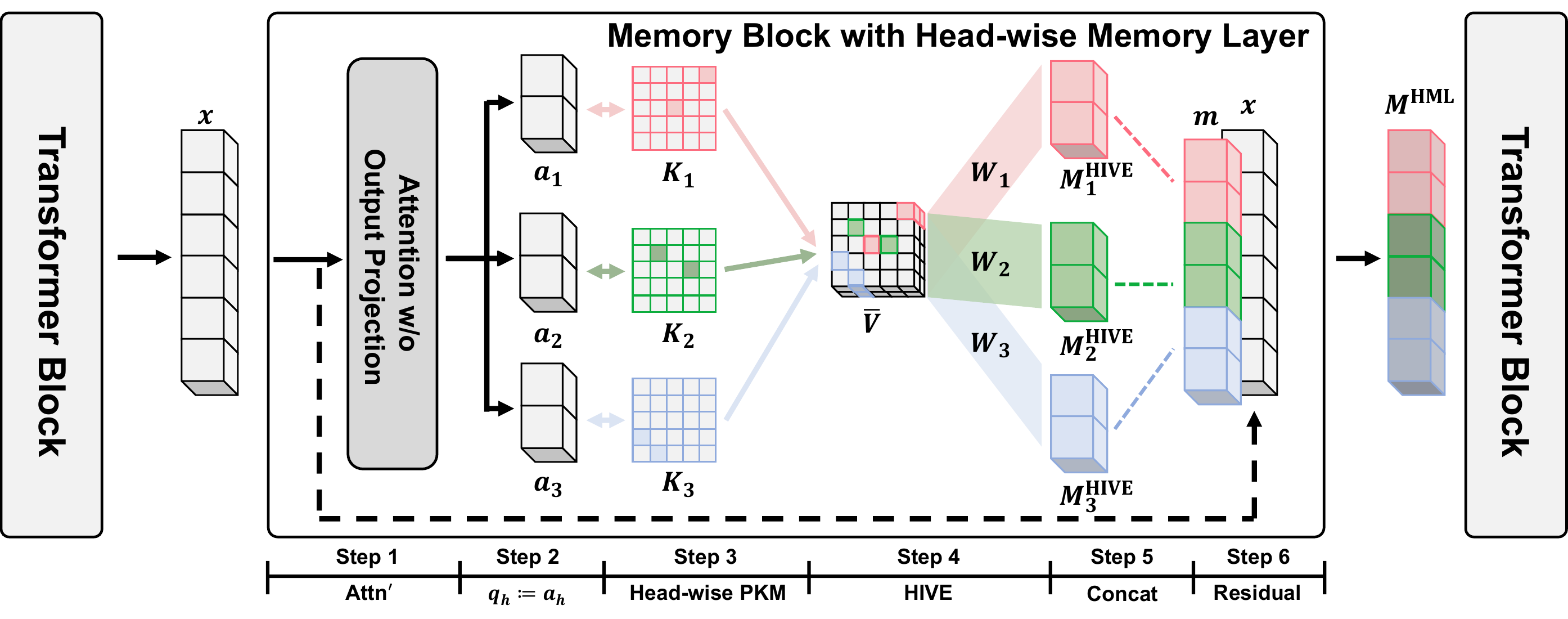}
    \caption{
MIDUS with HML consists of six steps, illustrated with $H=3$ attention heads
and Top-$k$ selection with $k=2$. 
1) The input $x$ entering the Memory block passes through
$\mathrm{Attn}^{\prime}$, an attention layer without output projection, producing
$a^{\prime}$. 
2) $a^{\prime}$ is split into head-wise representations $a_h$, which serve as
queries $q_h$ for each head. 
3) Each head uses its product-key set $K_h$ to retrieve the $k$ most similar
keys for $q_h$. 
4) The selected values from the shared latent value bank $\bar{V}$ are aggregated and
projected by the head-specific matrix $W_h$. 5) The head-wise outputs $M^{\mathrm{HIVE}}_{h}$ are concatenated to form $m$. 
6) Finally, $m$ is added to the input $x$ through a residual connection.
}
\label{fig:fig2}
    \vspace{-1em}
\end{figure*}

%% file: sec/theory.tex
We provide a structural analysis of added residual correction in MIDUS-HML under fixed retrieval coefficients. We focus on a head-wise correction setting in which heads require heterogeneous corrections within a common retrieval span. This setting isolates the value-side role of head-specific realization. We first compare a shared-realized-value family with a head-specific-value family, and then show that HIVE realizes the latter structure when the optimal head-specific banks admit a shared latent factorization. Under this condition, HIVE can recover the head-specific optimum, whereas a shared realized value bank incurs an irreducible gap when the projected targets differ across heads.

With fixed \(a'\) and retrieval coefficients, we collect the sparse Top-\(k\) retrieval weights \(\alpha_{h,t}\in\Delta^{N-1}\) into the matrix \(A_h=[\alpha_{h,1},\dots,\alpha_{h,T}]^\top\in\mathbb{R}^{T\times N}\) for each head \(h\). Let \(Y_h\in\mathbb{R}^{T\times d_h}\) denote the target correction for head \(h\), and let \(F(a')=[F_1(a')\mid\cdots\mid F_H(a')]\) be a head-partitioned correction. With \(\|\cdot\|\) denoting the Frobenius norm, we define the empirical loss as
\begin{equation}
\mathcal{L}(F)
=
\frac12\sum_{h=1}^{H}\|F_h(a')-Y_h\|^2. \nonumber
\end{equation}
We compare two value parameterizations under the same retrieval-weight matrices \(\{A_h\}_{h=1}^{H}\). The shared-realized-value family uses a single bank \(V_{\text{sh}}\in\mathbb{R}^{N\times d_h}\) with \(F_h(a')=A_hV_{\text{sh}}\), while the head-specific family uses banks \(V_h\in\mathbb{R}^{N\times d_h}\) with \(F_h(a')=A_hV_h\). Define
\[
\mathcal{L}_{\mathrm{Share}}^\star
=
\inf_{V_{\text{sh}}}\mathcal{L}(F),
\qquad
\mathcal{L}_{\mathrm{Ind}}^\star
=
\inf_{\{V_h\}_{h=1}^{H}}\mathcal{L}(F).
\]

\begin{restatable}{theorem}{structuredheterogeneity}
\label{thm:theory1}
In the fixed-retrieval setup, suppose \(A_h=A\) for all \(h\in[H]\), and let \(\Pi_A\) be the orthogonal projection onto the column space of \(A\). Assume that there exists an optimal head-specific solution \(\{V_h^\star\}_{h=1}^{H}\) that admits a shared latent factorization \(V_h^\star=\bar V W_h\), where \(\bar V\in\mathbb{R}^{N\times r}\), \(W_h\in\mathbb{R}^{r\times d_h}\), and \(r\le d_h\). If the projected targets are non-identical across heads, i.e., \(\Pi_AY_h\neq\Pi_AY_{h'}\) for some \(h\neq h'\), then the corresponding HIVE correction \(\widehat F_h(a')=A\bar V W_h\) satisfies
\[
\mathcal{L}(\widehat F)
=
\mathcal{L}_{\mathrm{Ind}}^\star
<
\mathcal{L}_{\mathrm{Share}}^\star.
\]
\end{restatable}

Proofs and additional propositions are provided in Appendix~\ref{app:theory}. The theorem clarifies the value-side mechanism behind the irreducible gap. HIVE shares storage through a single latent value table while realizing head-specific values through per-head projections, whereas a shared realized value bank must fit all heads with one value matrix. This supports the residual-correction view of MIDUS-HML, where shared value realization can be restrictive when heads share the same retrieval-weight matrix but require different projected corrections.

The result also connects our analysis to FFN-based DUS through the memory view of Transformer FFNs~\citep{geva2021transformer}. Under this view, an FFN can be viewed as using a single block-level value matrix to produce memory-like outputs over the full hidden representation. This provides an analogy to shared value realization, where duplicating FFN branches increases block-level value capacity but does not explicitly provide head-specific value realization. Our analysis is not an impossibility result for arbitrary FFNs. Instead, it isolates a fixed-retrieval regime where shared value realization is restrictive relative to head-specific value realization. HML with HIVE matches the latter structure by sharing a latent value table while allowing each attention head to realize its own residual correction.

%% file: sec/experiments.tex
\input{table/main_result.tex}

\subsection{Experiment Settings}
Following \citep{cao2025progressive}, we use Llama-3.2-1B and
Llama-3.1-8B~\citep{grattafiori2024llama} as base models, adding 8 and 16
blocks, respectively, for DUS and MIDUS. Unless specified,
each MIDUS memory layer uses a product-key bank with \(n=64\)
row/column sub-keys, i.e., \(N=n^2=4096\) composite keys per head, and
\(\mathrm{Top}\text{-}k\) retrieval with \(k=4\). Since both base models have 32 attention heads per block, replacing 8 blocks with HML yields 1.05M head-specific addressable memory slots, corresponding to product-key composite indices, for Llama-3.2-1B, while Llama-3.1-8B with 16 inserted blocks reaches 2.10M slots. For HIVE in HML, we set $r=d_h$. Boldface and underlining denote the best and second-best scores, respectively. Additional details and ablations are provided in the Appendix.

\paragraph{DUS placement policy.}
Block placement determines the finalized positions of added modules in the up-scaled model.
We use the placement definitions and final-stack positions summarized in
Appendix~\ref{appendix:policy}. DUS baselines follow their original placements, while
MIDUS uses the \textit{Distributed} placement by default. Given the selected final-stack
positions, MIDUS follows Eq.~\eqref{eq:midus}, where each Memory block is placed before the
corresponding pre-trained Transformer block and initialized from that block. We also
evaluate \textit{Top-heavy} and \textit{Bottom-heavy} MIDUS placements in Table~\ref{tab:dus_policy}.

\paragraph{CPT and SFT.}
We perform CPT on a 1.5B-token \texttt{FineWeb-Edu} subset
\citep{NEURIPS2024_370df50c,cao2025progressive} for general-purpose
language and reasoning, and on a 1.1B-token \texttt{MathPile} subset
\citep{wang2024mathpile} for math-domain adaptation. Following prior DUS
studies~\citep{wu2024llama,yang2025lesa,cao2025progressive}, we train
newly inserted blocks and freeze pre-existing base blocks. For the base
model and SOLAR~\citep{kim2023solar}, all parameters are updated. We
apply SFT to the \texttt{FineWeb-Edu} CPT model using
\texttt{Alpaca-GPT4}~\citep{peng2023instruction} and
\texttt{Databricks-Dolly-15k}~\citep{conover2023free} to evaluate
instruction-following transfer. During SFT, all parameters are updated,
following prior work \citep{yang2025lesa,cao2025progressive}.

\begin{figure}[t]
    \begin{minipage}[c]{0.54\textwidth}
        \input{table/efficient_8b}
    \end{minipage}%
    \hfill
    \begin{minipage}[c]{0.44\textwidth}
        \input{figure/prefill_8b}
    \end{minipage}
\vspace{-2em}
\end{figure}

\paragraph{Benchmarks.}
We evaluate models using the \texttt{lm-evaluation-harness} framework
\citep{gao2021framework}. For general-purpose language and reasoning,
we adopt the knowledge-centric suite used by \citep{cao2025progressive}:
ARC-Easy \citep{clark2018think}, LogiQA \citep{liu2020logiqa},
Winogrande \citep{sakaguchi2021winogrande}, CSQA
\citep{talmor2018commonsenseqa}, BoolQ \citep{clark2019boolq}, PIQA
\citep{bisk2020piqa}, and MMLU \citep{hendrycks2020measuring}, along
with perplexity (PPL) on WikiText \citep{merity2016pointer}. To assess
math-domain reasoning, we report performance on GSM8K, GSM8K--CoT
\citep{cobbe2021training}, MATH \citep{hendrycks2021measuring}, and
MathQA \citep{amini2019mathqa}. General-purpose benchmarks adopt
zero-shot evaluation, whereas math-domain benchmarks utilize 5-shot
evaluation. Additional evaluations on retrieval-heavy benchmarks,
including TriviaQA \citep{joshi2017triviaqa}, NQ-Open
\citep{lee2019latent}, and RACE \citep{lai2017race}, and long-context
evaluation on LongBench \citep{bai2024longbench} are provided in
Appendix~\ref{app:retrieval_long_context}.

\subsection{Experiment Results}
Table~\ref{tab:main_results} reports general-purpose language and reasoning performance after CPT on \texttt{FineWeb-Edu} and SFT on \texttt{Alpaca-GPT4}, using Llama-3.2-1B and Llama-3.1-8B backbones. Across both backbones, MIDUS-HML achieves the lowest WikiText PPL and the highest average zero-shot accuracy after CPT. These results indicate that replacing inserted FFN branches with HML provides an effective up-scaling strategy, where the added residual branch is parameterized as head-conditioned memory-based correction rather than dense FFN replication. The gains also carry over to SFT, suggesting that the CPT-trained MIDUS-HML model remains effective after instruction tuning. Appendix~\ref{appendix:dolby} reports a similar pattern on a different SFT dataset. We further examine retrieval-heavy and long-context benchmarks in Appendix~\ref{app:retrieval_long_context}, where MIDUS-HML also achieves the best average performance.

\input{./table/memory_performance}

Table~\ref{tab:memory_results} summarizes the CPT performance of MIDUS with different memory layers. For a fair comparison, all variants use the same number of memory heads, take per-head attention slices as queries, and maintain distinct key memories per head. Linear, PKM~\citep{lample2019large}, and UltraMem~\citep{huang2024ultra} rely on shared value parameterization across heads. Although UltraMem improves over baselines through virtual value, HML still achieves the best average accuracy, supporting the benefit of head-specific value realization through HIVE over shared value parameterization. Appendix~\ref{appendix:structure} provides further structural ablations of HML.

\paragraph{Efficiency of MIDUS-HML.}
Table~\ref{tab:efficiency_8b} reports parameter count, peak GPU memory,
training latency, and generation throughput for DUS and MIDUS variants
on Llama-3.1-8B. Among up-scaling methods, MIDUS-HML exhibits the
strongest efficiency profile. It uses the fewest trainable and total
parameters, achieves the lowest peak GPU memory in training and
inference, and outpaces the fastest DUS baseline, OpT-DeUS, in both
training time and generation throughput. Among memory-layer variants, HML is also the most efficient, using less GPU memory, achieving higher throughput, and training nearly twice as fast as Linear, PKM, and UltraMem. Figure~\ref{fig:prefill_8b}
analyzes prefill efficiency over prompt length. MIDUS-HML remains faster
than DUS across a broad range of prompt lengths, with the largest gains
at the longest prompts. While UltraMem expands the effective value space
through virtual value indexing and Tucker-based query--key scoring, this
design introduces larger prefill overhead as prompt length increases.
Note that all DUS methods share identical inference time efficiency and
therefore correspond to a single DUS reference curve in
Figure~\ref{fig:prefill_8b}.

DUS placement policy also affects training efficiency~\citep{cao2025progressive}.
Although our default MIDUS uses the Distributed policy, which is less efficiency-oriented than the Top-heavy policy used by OpT-DeUS, MIDUS-HML remains more efficient in most settings. With the same \textit{Top-heavy} policy, MIDUS-HML achieves higher training efficiency across batch sizes and sequence lengths, as shown in Figure~\ref{fig:abl_bs} and Figure~\ref{fig:abl_seq}. Further details are provided in Table~\ref{tab:dus_policy} and Appendix~\ref{appendix:policy}.

\input{table/math}
\input{figure/head_importance}

\paragraph{CPT results on \texttt{MathPile}.}
\texttt{MathPile} is compiled from diverse mathematical corpora~\citep{wang2024mathpile}, and we perform CPT with Llama-3.2-1B and Llama-3.1-8B backbones. Table~\ref{tab:math} reports 5-shot accuracy on math-domain benchmarks. Across both backbones, MIDUS-HML achieves the best average performance and remains competitive or better on most benchmarks. These results suggest that head-conditioned memory supports math-domain adaptation, including multi-step reasoning tasks.

\paragraph{Head-importance concentration under HML.}
Finally, to connect the empirical results with our head-wise motivation, we analyze whether HML is associated with more concentrated head-level contributions. To this end, we use the gradient-based head importance score \(IS_h\) proposed in prior work~\citep{bansal2023rethinking}. For examples $(x,y)\sim\mathcal{D}$, let $\ell(x,y)$ denote the autoregressive negative log-likelihood, and let $a_{h,t}([x\mid y])$ be the output of head $h$ at position $t$ in the sequence $[x\mid y]$. We define
\begin{equation}
IS_h(\mathcal{D})
:=
\mathbb{E}_{(x,y)\sim\mathcal{D}}
\left[
\left|
\sum_{t}
a_{h,t}([x\mid y])^\top
\frac{\partial \ell(x,y)}
{\partial a_{h,t}([x\mid y])}
\right|
\right].
\nonumber
\end{equation}
We compute $IS_h$ on PIQA in the zero-shot setting for the base Llama-3.2-1B model and MIDUS-HML after CPT on \texttt{FineWeb-Edu}. Although the base Transformer layers remain frozen, the interleaved HML layers change the head-importance profile. As shown in Figure~\ref{fig:head_importance}, MIDUS-HML produces a more uneven head-importance profile than the pre-trained base model. This increases the within-layer variance of \(IS_h\), suggesting that HML is associated with a more concentrated distribution of head-level contributions. This increased concentration is consistent with the view that HML allocates the added residual capacity in a head-conditioned manner, rather than applying a uniform expansion.

%% file: table/main_result.tex
\begin{table*}[t!]
\centering
\caption{Performance of DUS and MIDUS across Llama-3.2-1B and Llama-3.1-8B
backbones under CPT and SFT settings. Llama-3.1-8B$^\dagger$ results
are from \citep{cao2025progressive}, except for MIDUS-HML.}
\label{tab:main_results}
\vspace{-0.5em}
\begin{adjustbox}{width=\linewidth}
\setlength{\tabcolsep}{4pt}
\begin{tabular}{c c c c c c c c c c c c c c c c c c c c}
\toprule
& & \multicolumn{9}{c}{Llama-3.2-1B} & \multicolumn{9}{c}{Llama-3.1-8B$^\dagger$} \\ \cmidrule(lr){3-11}\cmidrule(lr){12-20}
& & \multicolumn{1}{c}{PPL $\downarrow$} & \multicolumn{8}{c}{Zero-shot Accuracy $\uparrow$} & \multicolumn{1}{c}{PPL $\downarrow$} & \multicolumn{8}{c}{Zero-shot Accuracy $\uparrow$} \\
\cmidrule(lr){3-3}\cmidrule(lr){4-11} \cmidrule(lr){12-12}\cmidrule(lr){13-20}
& Method & Wiki & ARC & LogiQA & Wino & CSQA & BoolQ & PIQA & MMLU & Average & Wiki & ARC & LogiQA & Wino & CSQA & BoolQ & PIQA & MMLU & Average \\
\midrule
\multirow{8}{*}{\rotatebox{90}{CPT}}
& Base & 13.22 & \textbf{68.69} & 22.27 & 60.22 & 25.88 & 63.61 & \underline{75.08} & 29.85 & 49.37 & 8.35 & 79.97 & 26.88 & 72.06 & 65.19 & 81.83 & 78.84 & 58.61 & 66.20 \\
& SOLAR & 13.41 & \underline{68.81} & 22.58 & 60.22 & 26.29 & 61.16 & \textbf{75.24} & 30.14 & 49.21 &  9.90 & 79.88 & 26.88 & 71.59 & 57.41 & 80.70 & 78.56 & 54.37 & 64.20 \\
& Llama Pro & 12.33 & 66.75 & 22.43 & 59.75 & 35.14 & \underline{64.86} & 74.54 & 31.22 & 50.67 &  7.81 & 81.61 & \textbf{29.49} & 73.72 & 70.93 & 81.65 & 79.98 & 62.56 & 68.56 \\
 & LESA   & 12.06 & 66.33 & 21.97 & 60.22 & 45.21 & 64.37 & 74.59 & 36.11 & 52.69 & \underline{7.73} & 82.07 & 27.96 & 74.11 & \textbf{72.40} & 81.93 & \underline{80.30} & 62.63 & 68.77 \\
& OpT-DeUS         & \underline{11.72} & 66.46 & \textbf{23.35} & \textbf{61.56} & \textbf{46.44} & 62.42 & 74.54 & \underline{36.56} & \underline{53.05} & \underline{7.73} & 82.07 & 27.34 & \textbf{74.74} & \underline{71.91} & \underline{82.26} & \textbf{80.79} & \underline{62.96} & \underline{68.87} \\
& Avg-DeUS         & 12.04 & 66.84 & 22.73 & \underline{60.38} & 41.93 & 64.01 & 74.27 & 34.40 & 52.08 &  7.95 & \underline{82.15} & 27.50 & 73.48 & 71.09 & 82.17 & 80.20 & 62.11 & 68.39 \\
\cmidrule(lr){2-20}
& MIDUS-HML & \textbf{11.64} & 66.16 & \underline{23.20} & \textbf{61.56} & \underline{46.27} & \textbf{65.29} & \underline{75.08} & \textbf{36.91} & \textbf{53.50} & \textbf{7.40} & \textbf{82.37} & \underline{28.57} & \underline{74.59} & 70.84 & \textbf{82.87} & 80.25 & \textbf{63.40} & \textbf{68.98} \\
\midrule
\multirow{7}{*}{\rotatebox{90}{SFT}}
& Base   & 13.09 & \textbf{69.65} & 21.20 & 58.96 & 26.37 & 62.66 & 75.46 & 30.43 & 49.25 & 8.32 & 81.10 & 24.58 & 72.14 & 68.30 & 82.14 & 79.71 & 59.17 & 66.73 \\
& SOLAR    &  13.23 & \underline{69.28} & \textbf{23.81} & 58.33 & 27.44 & 60.64 & \textbf{75.90} & 31.05 & 49.49 &  9.68 & 80.68 & 25.19 & 71.19 & 61.81 & 81.19 & 79.16 & 55.03 & 64.89 \\
& Llama Pro & 12.55 & 67.68 & 22.89 & 60.06 & 42.59 & 62.75 & 75.46 & 33.62 & 52.15 &  7.81 & 83.33 & \underline{27.19} & 74.11 & 72.07 & 82.26 & 80.79 & 62.32 & 68.87 \\
& LESA & 12.37 & 65.99 & 23.35 & \underline{60.38} & 47.91 & \textbf{66.33} & 75.24 & 36.43 & 53.66 &  \underline{7.72} & \underline{83.84} & 26.57 & \underline{75.53} & \underline{73.05} & 83.00 & 80.69 & 63.57 & 69.47 \\
& OpT-DeUS  & \underline{11.82} & 66.62 & \textbf{23.81} & 60.14 & \underline{49.55} & 63.43 & \underline{75.52} & \underline{37.64} & \underline{53.81} &  7.73 & 83.80 & 26.73 & \textbf{76.09} & \underline{73.05} & \underline{83.36} & \underline{80.85} & \underline{63.84} & \underline{69.67} \\
& Avg-DeUS  & 12.18 & 67.09 & \underline{23.66} & 60.30 & 47.01 & 65.63 & 74.97 & 35.62 & 53.47 &  7.91 & \textbf{83.88} & 26.42 & 75.45 & 72.89 & 83.18 & 80.47 & 63.10 & 69.34 \\
\cmidrule{2-11}
\cmidrule(lr){2-20}
& MIDUS-HML & \textbf{11.75} & 66.33 & 21.51 & \textbf{60.77} & \textbf{50.04} & \underline{66.21} & \textbf{75.90} & \textbf{37.82} & \textbf{54.08} & \textbf{7.50} & 83.50 & \textbf{28.11} & 74.90 & \textbf{73.63} & \textbf{83.39} & \textbf{80.96} & \textbf{64.54} & \textbf{69.86} \\
\bottomrule
\end{tabular}
\end{adjustbox}
\vspace{-1.7em}
\end{table*}

%% file: table/efficient_8b.tex
\centering
\captionof{table}{Efficiency for Llama-3.1-8B with the number of parameters (trainable/total), GPU memory (train/inference), training speed, and throughput. Base is measured under full-parameter training.}\label{tab:efficiency_8b}
\vspace{-0.5em}
\begin{adjustbox}{width=0.9\linewidth}
\setlength{\tabcolsep}{4pt}
\begin{tabular}{c c c c c c}
\toprule
& & \multicolumn{1}{c}{B $\downarrow$} & \multicolumn{1}{c}{GB $\downarrow$} & \multicolumn{1}{c}{s/iter $\downarrow$} & \multicolumn{1}{c}{Tokens/s $\uparrow$} \\
 \cmidrule(lr){3-3} \cmidrule(lr){4-4} \cmidrule(lr){5-5} \cmidrule(lr){6-6}
\multicolumn{2}{c}{Method} & Params & GPU Memory &  Train & Throughput \\
\midrule
& \textcolor{gray}{Base} & \textcolor{gray}{8.03 / 8.03} & \textcolor{gray}{81.6 / 19.2} & \textcolor{gray}{6.12} & \textcolor{gray}{61.0} \\
 \midrule \multirow{2}{*}{\rotatebox{90}{DUS}}
& Llama Pro & 3.49 / 11.52 & 63.8 / 27.0 & 6.89 & 41.3 \\
& OpT-DeUS & 3.49 / 11.52 & 59.8 / 27.0 & \underline{5.78}  & 41.3 \\
\midrule \multirow{3}{*}{\rotatebox{90}{MIDUS}}
& Linear & 1.48 / 9.51 & 51.3 / 23.1 & 11.05 & 42.4  \\
& PKM & \underline{1.21} / \underline{9.24} &  \underline{48.3} / \underline{22.5} & 9.88 & \underline{47.3} \\
& UltraMem & 1.29 / 9.32 &  56.9 / 23.6 & 8.56 & 36.9 \\
& HML & \textbf{0.42} / \textbf{8.45} & \textbf{34.8} / \textbf{20.1} & \textbf{5.05}  & \textbf{50.5} \\
\bottomrule
\end{tabular}
\end{adjustbox}

%% file: figure/prefill_8b.tex
\includegraphics[width=\linewidth]{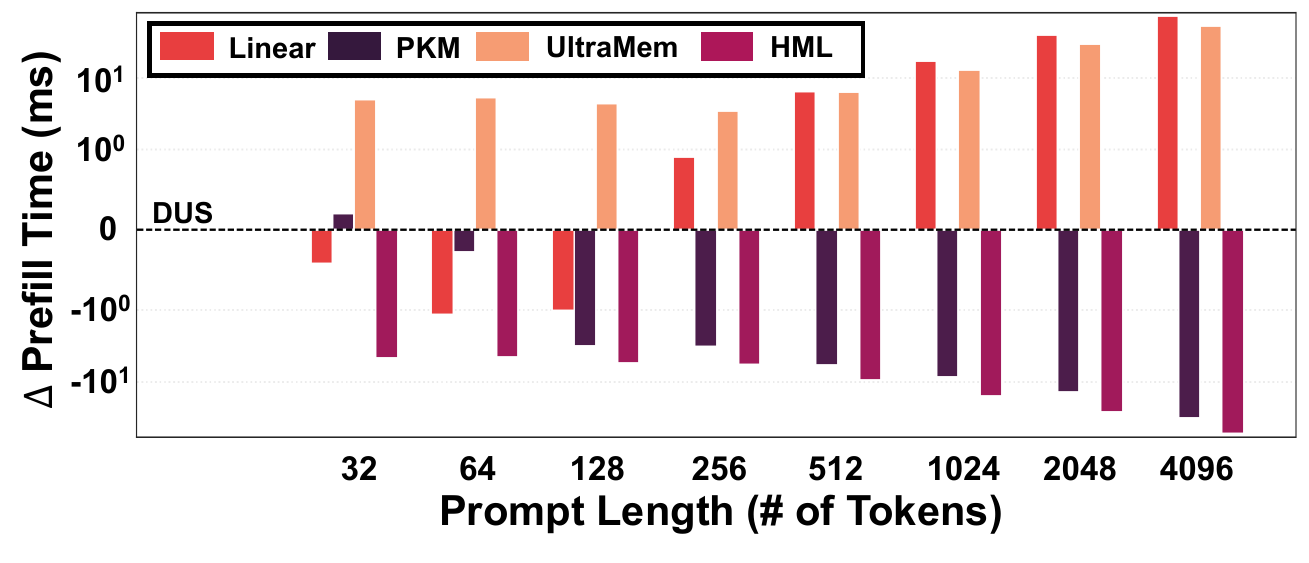}
\vspace{-1.5em}
\caption{Change in prefill time over prompt length for each memory layer relative to DUS, where negative $\Delta$ Prefill Time indicates faster prefill time than DUS. HML is the fastest across all prompt lengths.}
\label{fig:prefill_8b}

%% file: table/memory_performance.tex
\begin{wraptable}{r}{0.5\textwidth}
\centering
\vspace{-1.0em}
\caption{CPT results on \texttt{FineWeb-Edu} with Llama-3.2-1B for MIDUS variants with different memory layers. Linear, PKM, and UltraMem use the same number of memory heads as HML.}
\vspace{-0.5em}
\label{tab:memory_results}
\begin{adjustbox}{width=\linewidth}
\setlength{\tabcolsep}{4pt}
\begin{tabular}{c c c c c c c c c c}
\toprule
 & \multicolumn{1}{c}{PPL $\downarrow$} & \multicolumn{8}{c}{Zero-shot Accuracy $\uparrow$} \\
 \cmidrule(lr){2-2} \cmidrule(lr){3-10}
Mem & Wiki & ARC & LogiQA & Wino & CSQA & BoolQ & PIQA & MMLU & Average \\
\midrule 
Linear &  {\underline{11.82}} & 65.61 & \underline{22.73} & \textbf{62.04} & {42.42} & 64.71 & \underline{75.08} & {35.89} & 52.64 \\
PKM &  11.97 & 65.70 & 21.35 & 61.40 & 42.18 & \textbf{65.47} & {74.65} & 33.93 & 52.10 \\
UltraMem &  \textbf{11.64}  & \textbf{66.46} & 21.81  & 61.09 & \underline{44.31} & 64.83 & \textbf{75.63} & \underline{36.49} & \underline{52.94} \\
HML & \textbf{11.64} & \underline{66.16} & \textbf{23.20} & \underline{61.56} & \textbf{46.27} & \underline{65.29} & \underline{75.08} & \textbf{36.91} & \textbf{53.50} \\
\bottomrule
\end{tabular}
\end{adjustbox}
\vspace{-1em}
\end{wraptable}

%% file: table/math.tex
\begin{table*}[t!]
\centering
\caption{Performance of CPT on \texttt{MathPile} for Llama-3.2-1B and Llama-3.1-8B models.}
\label{tab:math}
\vspace{-0.5em}
\begin{adjustbox}{width=0.9\linewidth}
\begin{tabular}{c c c c c c c c c c c}
\toprule
& \multicolumn{5}{c}{Llama-3.2-1B (5-shot Accuracy $\uparrow$)}
& \multicolumn{5}{c}{Llama-3.1-8B (5-shot Accuracy $\uparrow$)} \\
\cmidrule(lr){2-6}\cmidrule(lr){7-11}
Method
& GSM8K & GSM8K--CoT & MATH & MathQA & Average
& GSM8K & GSM8K--CoT & MATH & MathQA & Average \\
\midrule
Base
& 4.32 & 4.93 & \underline{5.66} & 29.88 & 11.20
& 35.33 & 33.36 & 11.42 & 37.99 & 29.53 \\
Llama Pro
& 5.61 & 4.40 & 5.64 & 30.62 & 11.57
& 41.93 & 41.77 & 13.10 & \underline{40.87} & 34.42 \\
OpT-DeUS
& \underline{6.07} & \underline{6.22} & \textbf{5.80} & \underline{30.95} & \underline{12.26}
& \textbf{49.36} & \underline{50.04} & \underline{13.84} & \textbf{42.58} & \underline{38.96} \\ \midrule
MIDUS-HML
& \textbf{6.60} & \textbf{6.60} & 5.62 & \textbf{31.39} & \textbf{12.55}
& \underline{48.90} & \textbf{51.78} & \textbf{14.34} & \textbf{42.58} & \textbf{39.40} \\
\bottomrule
\end{tabular}
\end{adjustbox}
\vspace{-1em}
\end{table*}

%% file: figure/head_importance.tex
\begin{figure*}[t!]
    \centering
    \includegraphics[width=\linewidth]{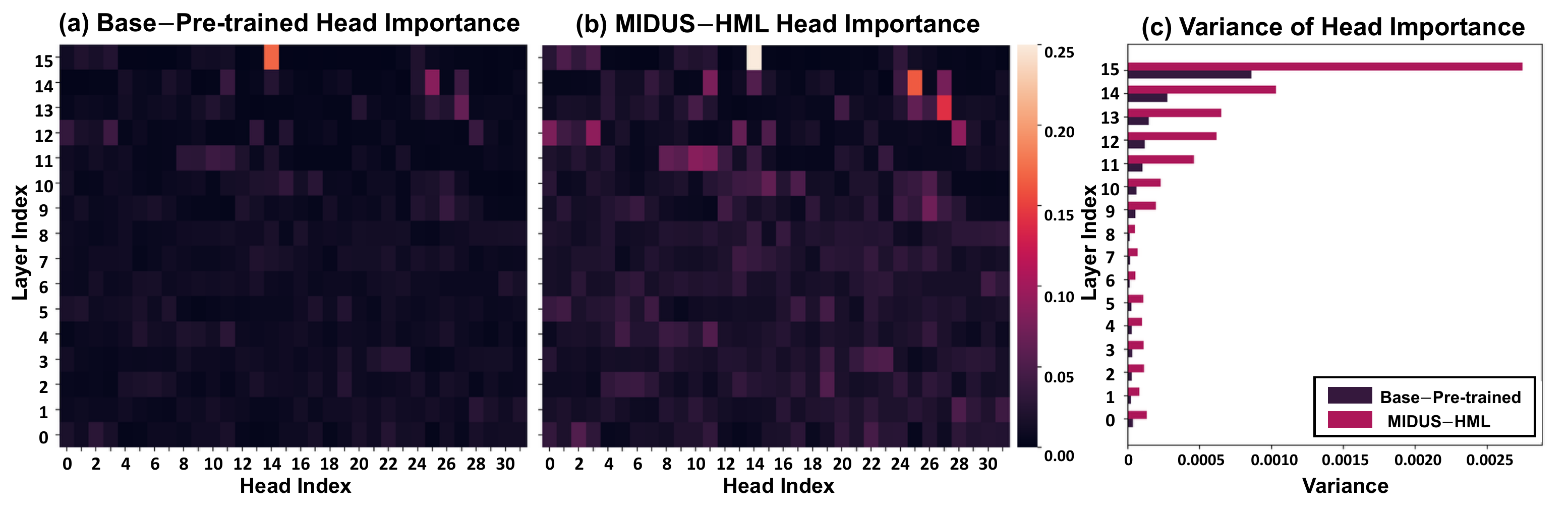}
    \vspace{-1.5em}
    \caption{Head-importance score $IS_h$ on PIQA for Llama-3.2-1B. (a) Heatmap of $IS_h$ over layer $\times$ head index for the pre-trained base model before CPT. (b) Heatmap for MIDUS-HML after CPT. Head importance rises overall, with the largest gains on heads already high in (a). (c) Per-layer variance of $IS_h$ across heads is consistently higher under MIDUS-HML, reflecting that important heads become more dominant. This pattern is consistent with head-conditioned capacity allocation in HML, where each head retrieves from its own key space and realizes head-specific values.}
    \label{fig:head_importance}
\vspace{-1.0em}
\end{figure*}

%% file: sec/conclusion.tex
We introduced MIDUS, a depth up-scaling framework that replaces duplicated FFN branches with memory layers, based on the intuition that added residual capacity in DUS need not remain tied to whole-block dense FFN replication. As a concrete instantiation, MIDUS-HML uses HML to allocate memory capacity at the attention-head level, where multi-head product-key memory enables head-wise retrieval and HIVE compactly realizes head-specific values from a shared latent bank. Across CPT and SFT settings, MIDUS-HML improves performance and efficiency over DUS baselines, using fewer parameters, lower GPU memory, and faster training and inference. Our head-importance and fixed-retrieval structural analyses further support head-wise memory as a practical and structurally distinct alternative to FFN-based residual expansion in DUS.

%% file: sec/appendix.tex
\clearpage
\section{Appendix}

\subsection{Limitations and Broader Impact.}
This work focuses on efficient DUS for Llama-3.2-1B and Llama-3.1-8B under CPT and SFT settings. While we evaluate MIDUS across general-purpose, math-domain, retrieval-heavy, and long-context benchmarks, further validation on other model families, larger scales, and different training regimes remains future work. Due to the computational cost of repeated CPT runs, our main results are reported as single-run point estimates rather than with error bars or confidence intervals. Our structural analysis should also be interpreted as a fixed-retrieval explanation of HIVE, not as a complete characterization of all FFN-based or memory-based expansion methods. MIDUS may reduce the parameter, memory, and computational costs of model up-scaling, improving accessibility and resource efficiency. However, more efficient LLM scaling may also lower the cost of deploying models that inherit broader LLM risks, including misuse, biased outputs, privacy concerns, and unsafe generations. These risks should be managed through the responsible-use policies and safeguards of the underlying models and deployment settings.

\subsection{Experimental Details} \label{appendix:details}

\subsubsection{Hyperparameter Settings}
We conduct \texttt{FineWeb-Edu} experiments on NVIDIA RTX 6000 Blackwell GPUs and \texttt{MathPile} experiments on NVIDIA H200 GPUs. Across all experiments, we use the AdamW optimizer \citep{loshchilov2017decoupled} with a cosine learning rate schedule, a warmup ratio of 0.1, and sweep weight decay in \{0, 1e-6, 1e-2\}. All models are trained in \texttt{bfloat16} with \texttt{FlashAttention-2} \citep{dao2023flashattention} enabled. For MIDUS-HML, to encourage balanced and sparse memory usage \citep{lample2019large} while limiting additional hyperparameter tuning, we fix the learning rates of key and value parameters to the maximum learning rate without scheduling and set their weight decay to zero. For Llama-3.2-1B on the \texttt{FineWeb-Edu} subset, we follow the hyperparameter configuration of prior work and apply the same search protocol to all baselines, including ours. During CPT, we search with a maximum learning rate of 1e-4, and during SFT, we fix the maximum learning rate to 1e-5. For Llama-3.1-8B on \texttt{FineWeb-Edu}, baseline results are taken from \citep{cao2025progressive}, while for MIDUS-HML we fix the CPT learning rate to 5e-5 and search the SFT learning rate over \{1e-6, 5e-6\}. In the \texttt{MathPile} experiments, we use a learning rate of 1e-4 for all methods and both backbones. Unless otherwise stated, for example in ablation studies, we fix the global batch size at 64 and the sequence length at 2{,}048 tokens.

\subsubsection{Dataset Details}
Following \citep{cao2025progressive}, we construct a 1.5B-token \texttt{FineWeb-Edu} subset by sampling from the CC-MAIN-2024-51 slice of \texttt{FineWeb-Edu} \citep{NEURIPS2024_370df50c} after the base model's pre-training cut-off. For SFT, we use the instruction-tuning datasets \texttt{Alpaca-GPT4} (52k examples) \citep{peng2023instruction} and \texttt{Databricks-Dolly-15k} (15k examples) \citep{conover2023free}. To construct the \texttt{MathPile} subset, we extract math-related text from the five components of the original \textsc{MathPile} \citep{wang2024mathpile} corpus (Textbooks, Wikipedia, ProofWiki, CommonCrawl, StackExchange), excluding arXiv, resulting in approximately 1.1B tokens.

\subsection{Implementation Details} \label{appendix:implement}

MIDUS preserves the interleaved depth-expansion structure of DUS, but replacing duplicated FFN branches with HML changes the computational profile of the inserted blocks. Unlike FFNs, which are dominated by regular matrix multiplications (GEMMs), HML involves sparse retrieval with discrete addressing, per-sample weighting, and index-based accumulation. Existing deep learning frameworks, such as PyTorch, provide highly optimized kernels for GEMM-heavy workloads, but sparse lookup and weighted accumulation are less directly optimized. As a result, na\"ive HML implementations may suffer from irregular memory access, limited hardware utilization, and atomic operation overhead in CUDA kernels.

In this section, we describe three implementation-level optimizations used for HML to reduce the gap between theoretical and practical efficiency. These optimizations are orthogonal to the architectural design of HML and operate at the kernel and execution level. Further low-level engineering, such as fully fused CUDA kernels for the entire HML block, may yield additional gains beyond those reported in this work.

\subsubsection{Fused Cartesian Top-k for Efficient Generation}
\label{app:fused_topk}

The head-wise product-key selection used in HML avoids scoring all $N=n^2$ composite keys by using a two-stage selection procedure. As described in Section~\ref{sec:hml}, for each head $h$ and token $t$, the model first selects Top-$k$ row and column index sets $I_{h,t}$ and $J_{h,t}$, forms the candidate pair set $\widehat{\Omega}_{h,t}=I_{h,t}\times J_{h,t}$, and then selects the final Top-$k$ pair set $\Omega_{h,t}$ using the additive pair scores $\sigma_{h,t}$. This row--column factorization reduces the pair-selection work from scoring all $N=n^2$ composite keys to scoring two $n$-way sub-key sets per head. In practice, however, this hierarchical procedure can introduce non-trivial kernel-launch overhead and irregular memory access on GPUs, especially when the number of query tokens and the batch size are small, as in autoregressive decoding.

To reduce this overhead, we use a \emph{Fused Cartesian Top-$k$} strategy for short-query inference. While the main HML formulation uses the two-stage head-wise product-key selection, the inference path optionally replaces this hierarchical selection with a fused operation. For each head $h$ and token $t$, we construct the Cartesian score grid $\mathcal{S}_{h,t}\in\mathbb{R}^{n\times n}$ by
\begin{equation}
    (\mathcal{S}_{h,t})_{ij}
    =
    S^{\mathrm{row}}_{h,t}(i)
    +
    S^{\mathrm{col}}_{h,t}(j).
    \nonumber
\end{equation}
We then let $\Omega_{h,t}\subset\{1,\dots,n\}\times\{1,\dots,n\}$ be the Top-$k$ pair set over the flattened grid $\mathcal{S}_{h,t}$. This is equivalent to scoring all $n^2$ product-key pairs, but it is executed as a single fused operation that better utilizes GPU parallelism.

This fused strategy sacrifices the asymptotic $\mathcal{O}(n)$ pair-selection cost and uses $\mathcal{O}(N)$ work per head and query token. However, $n$ is fixed and modest in our setting, and generation typically uses few query tokens or small batches. In this regime, reduced kernel-launch overhead and improved hardware utilization can lower end-to-end latency. For longer sequences and larger batches, we use the standard two-stage product-key selection described in Section~\ref{sec:hml}.

\subsubsection{Head-wise Value Caching for Inference}
\label{app:value_caching}

During training, HIVE retrieves latent values from the shared latent value bank $\bar V$ and applies a head-specific value projection $W_h$, as defined in Eq.~\eqref{eq:hive}:
\begin{equation}
    \bar M_{h,t}
    =
    \sum_{(i,j)\in\Omega_{h,t}}
    \alpha_{h,t}(i,j)\,
    \bar V_{\pi(i,j)}
    \in\mathbb{R}^{r},
    \qquad
    M^{\mathrm{HIVE}}_{h,t}
    =
    \bar M_{h,t} W_h
    \in\mathbb{R}^{d_h}.
    \nonumber
\end{equation}
This factorization shares latent value storage while allowing head-specific value realization, reducing the training-time parameter cost compared with maintaining an explicit value table of size $N\times d_h$ for each head.

During inference, model weights are fixed, so repeatedly applying $W_h$ to retrieved latent values is redundant. We remove this overhead by constructing head-specific cached value tables only for inference. Before decoding, for each head $h$, we compute
\begin{equation}
    V^{\mathrm{cache}}_h
    =
    \bar V W_h
    \in \mathbb{R}^{N \times d_h}.
    \nonumber
\end{equation}
This is equivalent to applying the head-specific value projection to every row of $\bar V$. At runtime, the memory layer retrieves from $V^{\mathrm{cache}}_h$ using the selected pair indices and weights, yielding the same output as applying $W_h$ after latent-value aggregation. This equivalence follows from the linearity of the value projection.

This optimization trades additional inference-time storage for lower decoding latency. It does not change the training-time parameterization or storage footprint of HIVE. The learned parameters remain factorized as $(\bar V,\{W_h\}_{h=1}^{H})$, and the cached tables are constructed on demand only for inference. All inference-side efficiency metrics reported in our experiments, including peak GPU memory during decoding, generation throughput, and prefill time, are measured with value caching enabled.

\subsubsection{Atomic Contention Mitigation via Deduplicated Gradient Accumulation}
\label{app:grad_conflict}

A practical bottleneck in training memory layers arises in the backward pass of value aggregation. When multiple retrieval queries access the same value-table row, standard implementations such as \texttt{torch.nn.EmbeddingBag} or \texttt{scatter\_add} may issue many concurrent updates to the same memory address. On GPUs, these updates are typically serialized through atomic operations, which can reduce effective memory bandwidth and slow training.

This issue is particularly relevant for our memory layers, which combine sparse retrieval, per-sample weights, and \texttt{bfloat16} values. Current PyTorch kernels do not provide an efficient backward path for this combination that matches our requirements. We therefore implement a custom autograd function based on deduplication and pre-aggregation. The scheme applies to any value aggregation of the form
\begin{equation}
y_p=\sum_{u=1}^{k}\lambda_{p,u}V_{\mathcal{I}_{p,u}},
\qquad p=1,\dots,P,
\nonumber
\end{equation}
where \(P\) is the number of retrieval queries, \(k\) is the number of retrieved entries per query, \(\mathcal{I}_{p,u}\in\{1,\dots,N\}\) is a value-table index, and \(\lambda_{p,u}\) is the corresponding retrieval weight.

Given the output gradient \(G_Y\), the gradient for value row \(j\) is
\begin{equation}
G_V[j]
=
\sum_{(p,u):\,\mathcal{I}_{p,u}=j}
\lambda_{p,u}G_Y[p].
\nonumber
\end{equation}
Algorithm~\ref{alg:backward} computes this gradient by first forming per-retrieval contributions, then deduplicating value indices, and finally aggregating all contributions for each unique value row before writing to the global gradient table. This reduces repeated atomic updates to the same global memory location.

\begin{algorithm}[h]
\small
\caption{Deduplicated Backward Pass for Value Aggregation}
\label{alg:backward}
\begin{algorithmic}[1]
\REQUIRE Output gradient \(G_Y \in \mathbb{R}^{P \times d_v}\), value indices \(\mathcal{I} \in \{1,\dots,N\}^{P \times k}\), retrieval weights \(\Lambda \in \mathbb{R}^{P \times k}\), value-table size \(N\)
\STATE \textbf{Step 1: Form per-retrieval gradient contributions}
\STATE \(G_{\mathrm{exp}} \leftarrow G_Y.\mathrm{unsqueeze}(1) \odot \Lambda.\mathrm{unsqueeze}(-1)\) \COMMENT{\(G_{\mathrm{exp}}\in\mathbb{R}^{P\times k\times d_v}\)}
\STATE \(G_{\mathrm{flat}} \leftarrow G_{\mathrm{exp}}.\mathrm{reshape}(Pk,d_v)\)
\STATE \(\mathcal{I}_{\mathrm{flat}} \leftarrow \mathcal{I}.\mathrm{reshape}(Pk)\)
\STATE \textbf{Step 2: Deduplicate retrieved value indices}
\STATE \(\mathcal{I}_{\mathrm{uniq}}, \eta \leftarrow \mathrm{unique}(\mathcal{I}_{\mathrm{flat}}, \mathrm{return\_inverse=True})\)
\STATE \textbf{Step 3: Pre-aggregate contributions for each unique index}
\STATE \(U \leftarrow |\mathcal{I}_{\mathrm{uniq}}|\)
\STATE \(G_{\mathrm{uniq}} \leftarrow \mathrm{zeros}(U,d_v)\)
\STATE \(G_{\mathrm{uniq}}.\mathrm{index\_add\_}(0,\eta,G_{\mathrm{flat}})\)
\STATE \textbf{Step 4: Write one aggregated gradient per touched value row}
\STATE \(G_V \leftarrow \mathrm{zeros}(N,d_v)\)
\STATE \(G_V[\mathcal{I}_{\mathrm{uniq}}] \leftarrow G_{\mathrm{uniq}}\)
\ENSURE \(G_V\)
\end{algorithmic}
\end{algorithm}

This backward scheme computes the same value-table gradient as the naive implementation, up to floating-point accumulation order, while reducing repeated global atomic updates to the same value-table row. Gradients with respect to the retrieval weights are computed separately following the standard weighted-embedding rule,
\[
\frac{\partial \mathcal{L}}{\partial \lambda_{p,u}}
=
\left\langle G_Y[p], V_{\mathcal{I}_{p,u}}\right\rangle,
\]
and are omitted from Algorithm~\ref{alg:backward} for brevity.

\subsection{DUS Placement Policy} \label{appendix:policy}

DUS placement policy is a key design choice that affects both CPT performance and training efficiency. The Llama-3.2-1B and Llama-3.1-8B base models consist of 16 and 32 Transformer layers, respectively. For all DUS and MIDUS variants, we insert 8 and 16 additional modules into these backbones, yielding final depths of 24 and 48 blocks.

We distinguish between the underlying anchor schedule and the final-stack positions of inserted modules. DUS and MIDUS can use the same anchor schedule, but their inserted modules appear at different final-stack indices because they follow different insertion rules. In standard interleaved DUS, as in Eq.~\eqref{eq:dus}, each duplicated Transformer block is placed after its anchor block. In MIDUS, as in Eq.~\eqref{eq:midus}, each Memory block is placed before the corresponding pre-trained block. Therefore, the \textit{Llama Pro} placement used for DUS baselines and the \textit{Distributed} placement used for MIDUS correspond to the same underlying anchor schedule, but their inserted-module indices are shifted in the final up-scaled stack. Although \textit{Llama Pro} and \textit{Distributed} have different final-stack indices, they correspond to the same underlying anchor schedule under different insertion rules.

We consider four final-stack placement patterns:
\begin{itemize}
    \item \textit{Top-heavy}: inserted modules are concentrated in the upper half of the stack, alternating with pre-trained blocks, following the recipe of~\citep{cao2025progressive}.
    \item \textit{Llama Pro}: inserted Transformer blocks are placed every three positions starting from index 2 in the final stack, following the DUS insertion rule of~\citep{wu2024llama}.
    \item \textit{Distributed}: Memory blocks are placed every three positions starting from index 1 in the final stack. This is the MIDUS counterpart of the Llama Pro anchor schedule, shifted by the before-anchor insertion rule in Eq.~\eqref{eq:midus}.
    \item \textit{Bottom-heavy}: inserted modules are concentrated in the lower half of the stack, alternating with pre-trained blocks.
\end{itemize}

The following lists report the resulting positions of the inserted modules in the final up-scaled stack, using 0-indexing. For DUS baselines, the listed indices indicate the positions of duplicated Transformer blocks. For MIDUS variants, the listed indices indicate the positions of Memory blocks, with the corresponding pre-trained block placed immediately after each Memory block under Eq.~\eqref{eq:midus}. Thus, although \textit{Llama Pro} and \textit{Distributed} have different final-stack indices, they instantiate the same underlying anchor schedule under different insertion rules.

\paragraph{Up-scaled Llama-3.2-1B with 24 blocks}
\begin{itemize}
    \small
    \item \textit{Top-heavy}: $\{8,\,10,\,12,\,14,\,16,\,18,\,20,\,22\}$
    \item \textit{Llama Pro}: $\{2,\,5,\,8,\,11,\,14,\,17,\,20,\,23\}$
    \item \textit{Distributed}: $\{1,\,4,\,7,\,10,\,13,\,16,\,19,\,22\}$
    \item \textit{Bottom-heavy}: $\{0,\,2,\,4,\,6,\,8,\,10,\,12,\,14\}$
\end{itemize}

\paragraph{Up-scaled Llama-3.1-8B with 48 blocks}
\begin{itemize}
    \small
    \item \textit{Top-heavy}: $\{16,\,18,\,20,\,22,\,24,\,26,\,28,\,30,\,32,\,34,\,36,\,38,\,40,\,42,\,44,\,46\}$
    \item \textit{Llama Pro}: $\{2,\,5,\,8,\,11,\,14,\,17,\,20,\,23,\,26,\,29,\,32,\,35,\,38,\,41,\,44,\,47\}$
    \item \textit{Distributed}: $\{1,\,4,\,7,\,10,\,13,\,16,\,19,\,22,\,25,\,28,\,31,\,34,\,37,\,40,\,43,\,46\}$
    \item \textit{Bottom-heavy}: $\{0,\,2,\,4,\,6,\,8,\,10,\,12,\,14,\,16,\,18,\,20,\,22,\,24,\,26,\,28,\,30\}$
\end{itemize}

When considering only the placement policy, \textit{Top-heavy} is the most training-efficient design under our expanded-module training setup, since placing inserted modules closer to the output reduces the backward computation and activation memory associated with these modules. \textit{Llama Pro} is also more efficient than \textit{Distributed} placement for the same reason. In contrast, MIDUS-HML achieves its best performance under the \textit{Distributed} policy, yet still provides substantially better overall efficiency than DUS baselines. MIDUS-HML can also adopt the \textit{Top-heavy} policy to further improve efficiency; even in this extreme setting, its performance remains comparable to, or better than, that of DUS baselines. See Table~\ref{tab:dus_policy} and Appendix~\ref{appendix:efficient} for further ablation studies.

\clearpage
\subsection{SFT results on \texttt{Databricks-Dolly-15k}}\label{appendix:dolby}

\input{./table/1b_dolby}

\subsection{Additional Evaluation on Retrieval-Heavy and Long-Context Benchmarks}
\label{app:retrieval_long_context}

To test whether the gains of MIDUS-HML extend beyond the main knowledge-centric benchmark suite, we further evaluate retrieval-heavy and long-context tasks. We use the same CPT-trained Llama-3.2-1B models from the FineWeb-Edu experiments. For retrieval-heavy evaluation, we use TriviaQA~\citep{joshi2017triviaqa}, NQ-Open~\citep{lee2019latent}, and RACE~\citep{lai2017race} under 5-shot prompting. For long-context evaluation, we use LongBench~\citep{bai2024longbench} and report results over six categories.

\input{./table/retri_heavy}
\input{./table/long_context}

\subsection{Ablation Studies} \label{appendix:ablation}
\subsubsection{Ablation study on the number of memory slots}
\input{table/num_memory}
In Table~\ref{tab:memory_count}, we vary the total number of product-key memories from tens of thousands to one million, i.e., $n{=}16,32,64$ in the PKM factorization. Average zero-shot accuracy rises steadily with capacity, while PPL stays essentially flat. We hypothesize that a larger number of memory slots mainly helps tasks that benefit from retrieving facts or patterns, without disrupting the core language-modeling behavior. MIDUS gains an additional axis to scale quality, and rather than adding more Transformer blocks, we can improve results by increasing the number of memory slots.

\subsubsection{Ablation study on Top-k}
In Table~\ref{tab:topk}, we vary the number of retrieved memory slots per token from $k{=}1$ to $k{=}8$ while keeping the memory size and all other hyperparameters fixed. Average zero-shot accuracy improves steadily as $k$ increases from 1 to 4, with modest or no further gains at $k{=}8$, and PPL remains essentially unchanged across all settings. We therefore use $k{=}4$ as the default Top-$k$ setting in all other experiments.
\input{table/topk}

\subsubsection{Ablation study on the learning rate and weight decay}
\input{./table/hyperparam}
In Table~\ref{tab:lr_wd}, we vary the learning rate and weight decay over a range of reasonable values while keeping all other settings fixed. Across these configurations, PPL remains tightly clustered around 11.62–11.64 and the average zero-shot accuracy stays in a narrow band, indicating that MIDUS-HML is relatively robust to these optimization hyperparameters.

\subsubsection{Ablation study on the Structure of HML}
\label{appendix:structure}

\input{table/ablation}
In Table~\ref{tab:ablations}, we evaluate four HML variants. The first applies BatchNorm1d to the queries, following PKM~\citep{lample2019large}. The second applies LayerNorm to the queries. The third adds an internal attention residual, \(a' = x + \mathrm{Attn}'(\mathrm{LN}(x))\). The fourth enables the attention output projection. Neither query-normalized variant improves over the default HML, suggesting that normalization may disturb the score structure used for Top-$k$ selection. The internal residual also degrades performance, which indicates that directly using separated head outputs may provide more suitable memory queries than mixing them with the input representation. Although adding the output projection gives a marginal gain, we retain the projection-free design as the default because it preserves head-separated queries and avoids the additional projection cost.

\subsubsection{Ablation study on DUS placement policy.}
\input{table/ablation_freeze}
Table~\ref{tab:dus_policy} shows the efficiency and performance of MIDUS-HML with Llama-3.2-1B under different DUS placement policies. \textit{Top-heavy} placement, as used by OpT-DeUS \citep{cao2025progressive}, is most efficient in memory and time, since many frozen lower blocks can be skipped during backpropagation. \textit{Bottom-heavy} placement slightly improves PPL but does not maximize average zero-shot accuracy. \textit{Distributed} policy yields the best overall accuracy.


\subsection{Additional efficiency experiments}
\label{appendix:efficient}
\input{./figure/abl_bs}
\input{./figure/abl_seq}
We measure all efficiency metrics on a single NVIDIA RTX 6000 Blackwell GPU. For training-related metrics, Table~\ref{tab:efficiency_8b} and Figure~\ref{fig:prefill_8b} use a global batch size of 16 with 16 gradient-accumulation steps. Unless otherwise noted, the sequence length is fixed to 2048 in these efficiency experiments. For inference-related metrics, we report peak GPU memory and generation throughput measured during the generation stage.

Figure~\ref{fig:abl_bs} further examines efficiency as a function of the global batch size. We fix the gradient-accumulation steps to 16 and the sequence length to 2048, and vary only the global batch size. For Llama-3.1-8B, we omit the training-time entries for Llama Pro and OpT-DeUS at a global batch size of 64, since they exceed the available GPU memory on a single RTX 6000 Blackwell GPU. Across Figure~\ref{fig:abl_bs} and \ref{fig:abl_seq}, which cover a range of batch sizes and sequence lengths, MIDUS-HML (\textit{Dist.}) consistently achieves training times that are faster than or comparable to those of the DUS baselines. Notably, MIDUS-HML (\textit{Top.}), which adopts the same DUS placement policy as OpT-DeUS, is substantially faster than OpT-DeUS. In all cases, MIDUS-HML also requires less peak GPU memory than any of the DUS baselines. The Llama-3.1-8B results suggest that the efficiency advantage of MIDUS-HML remains pronounced on a larger and deeper backbone.

\input{./sec/theory_appendix}

%% file: table/1b_dolby.tex
\begin{table*}[h!]
\centering
\caption{SFT results on the \texttt{Databricks-Dolly-15k} \citep{conover2023free} with Llama-3.2-1B.}
\label{tab:sft_dolly}
\begin{adjustbox}{width=0.60\linewidth}
\setlength{\tabcolsep}{4pt}
\begin{tabular}{c c c c c c c c c c c}
\toprule
& & \multicolumn{9}{c}{Llama-3.2-1B} \\ \cmidrule{3-11}
& & \multicolumn{1}{c}{PPL $\downarrow$} & \multicolumn{8}{c}{Zero-shot Accuracy $\uparrow$} \\
\cmidrule(lr){3-3}\cmidrule(lr){4-11}
& Method & Wiki & ARC & LogiQA & Wino & CSQA & BoolQ & PIQA & MMLU & Average \\
\midrule
\multirow{7}{*}{\rotatebox{90}{SFT}}
& Base & 13.07 & \underline{68.64} & 21.35 & 60.30 & 27.52 & 63.27 & \textbf{75.24} & 30.49 & 49.55 \\
& SOLAR & 13.19 & \textbf{69.40} & 23.04 & 59.04 & 27.52 & 60.03 & 75.14 & 30.77 & 49.28 \\
& Llama Pro & 12.47 & 68.10 & \textbf{23.81} & 60.54 & 41.93 & 63.70 & \textbf{75.24} & 34.03 & 52.48 \\
& LESA   & \underline{11.77} & 66.46 & \underline{23.66} & 60.77 & \underline{49.06} & 64.98 & 74.86 & \underline{38.22} & 54.00 \\
& OpT-DeUS  & 11.78 & 67.21 & \textbf{23.81} & \underline{61.48} & \textbf{49.14} & 63.76 & \underline{75.19} & 37.83 & \underline{54.06} \\
& Avg-DeUS   & 12.10 & 67.51 & 22.73 & 60.30 & 46.60 & \underline{65.02} & 74.37 & 36.79 & 53.33 \\
\cmidrule(lr){2-11}
& MIDUS-HML & \textbf{11.72} & 66.62 & 22.12 & \textbf{61.72} & 48.65 & \textbf{66.51} & 75.14 & \textbf{38.59} & \textbf{54.19} \\
\bottomrule
\end{tabular}
\end{adjustbox}
\end{table*}

%% file: table/retri_heavy.tex
\begin{table}[h!]
\centering
\caption{Retrieval-heavy evaluation of CPT-trained Llama-3.2-1B models.}
\label{tab:retrieval_heavy}
\begin{adjustbox}{width=0.45\linewidth}
\begin{tabular}{ccccc}
\toprule
Method & TriviaQA & NQ-Open & RACE & Average \\
\midrule
Base      & 30.38 & 9.72  & \textbf{38.28} & 26.13 \\
Llama Pro & 34.33 & 10.28 & 37.99 & 27.53 \\
OpT-DeUS  & \underline{39.02} & \underline{12.11} & \underline{38.18} & \underline{29.77} \\
Avg-DeUS  & 38.25 & 11.75 & 37.42 & 29.14 \\
MIDUS-HML & \textbf{39.75} & \textbf{12.24} & 37.51 & \textbf{29.83} \\
\bottomrule
\end{tabular}
\end{adjustbox}
\end{table}

%% file: table/long_context.tex
\begin{table}[h!]
\centering
\caption{Long-context evaluation of CPT-trained Llama-3.2-1B models on LongBench. Overall denotes the task-level average over the six categories.}
\label{tab:long_context}
\begin{adjustbox}{width=0.8\linewidth}
\begin{tabular}{cccccccc}
\toprule
Method & Single-Doc QA & Multi-Doc QA & Summarization & Few-shot & Synthetic & Code & Overall \\
\midrule
Base      & 10.45 & 7.40 & 19.10 & 55.68 & \textbf{4.15} & 27.47 & 21.32 \\
Llama Pro & 10.52 & 7.32 & 21.09 & 58.08 & 3.35 & 35.70 & 23.07 \\
OpT-DeUS  & \textbf{10.92} & \underline{7.97} & \underline{22.94} & \underline{60.60} & \underline{3.79} & \underline{47.57} & \underline{25.63} \\
Avg-DeUS  & \underline{10.66} & 7.74 & 21.04 & 60.38 & 3.54 & 46.96 & 25.03 \\
MIDUS-HML & 10.60 & \textbf{8.48} & \textbf{23.23} & \textbf{63.52} & 2.62 & \textbf{54.64} & \textbf{27.00} \\
\bottomrule
\end{tabular}
\end{adjustbox}
\end{table}

%% file: table/num_memory.tex
\begin{table}[h!]
\centering
\caption{Ablation study on the number of total memory slots and its effect on performance. The 66K, 262K, and 1M settings denote total addressable slots across inserted HML blocks and memory heads; each head uses \(N=n^2\) composite keys.}
\label{tab:memory_count}
\begin{adjustbox}{width=0.6\linewidth}
\setlength{\tabcolsep}{4pt}
\begin{tabular}{c c c c c c c c c c}
\toprule
& \multicolumn{1}{c}{PPL $\downarrow$} & \multicolumn{8}{c}{Zero-shot Accuracy $\uparrow$} \\
\cmidrule(lr){2-2}\cmidrule(lr){3-10}
\# Memories & Wiki & ARC & LogiQA & Wino & CSQA & BoolQ & PIQA & MMLU & Average \\
\midrule
66K ($n=16$) & \textbf{11.63} & \underline{66.12} & 22.89 & 60.06 & \underline{46.27} & 65.02 & \underline{74.81} & \underline{36.74} & 53.13 \\
262K ($n=32$) &  \underline{11.64} & 65.95 & \textbf{23.35} & \underline{60.85} & \textbf{47.17} & \textbf{65.44} & 74.76 & 36.41 & \underline{53.42} \\
1M ($n=64$) &  \underline{11.64} & \textbf{66.16} & \underline{23.20} & \textbf{61.56} & \underline{46.27} & \underline{65.29} & \textbf{75.08} & \textbf{36.91} & \textbf{53.50} \\
\bottomrule
\end{tabular}
\end{adjustbox}
\end{table}

%% file: table/topk.tex
\begin{table}[h!]
\centering
\caption{Effect of the Top-$k$ value $k$ for Llama-3.2-1B with MIDUS-HML.}
\label{tab:topk}
\begin{adjustbox}{width=0.7\linewidth}
\setlength{\tabcolsep}{4pt}
\begin{tabular}{c c c c c c c c c c c c}
\toprule
& \multicolumn{1}{c}{Train Time $\downarrow$} & \multicolumn{1}{c}{TTFT $\downarrow$} & \multicolumn{1}{c}{PPL $\downarrow$} & \multicolumn{8}{c}{Zero-shot Accuracy $\uparrow$} \\
\cmidrule(lr){2-2}\cmidrule(lr){3-3}\cmidrule(lr){4-4}\cmidrule(lr){5-12}
Top-$k$ & s/iter & ms & Wiki & ARC & LogiQA & Wino & CSQA & BoolQ & PIQA & MMLU & Average \\
\midrule
1 & \textbf{4.31} & \underline{9.33} & 11.67 & 65.74 & 21.97 & 60.54 & 46.19 & 64.95 & 74.86 & 36.95 & 53.03 \\
2 & \underline{4.41} & 9.40 &\textbf{11.63} & 65.87 & \textbf{23.20} & \underline{60.69} & 46.11 & \underline{65.11} & 74.54 & \underline{37.04} & 53.22 \\
4 & 4.54 & \textbf{9.31} &\underline{11.64} & \textbf{66.16} & \textbf{23.20} & \textbf{61.56} & \underline{46.27} & \textbf{65.29} & \textbf{75.08} & 36.91 & \textbf{53.50} \\
8 & 4.81 & 10.71 & \underline{11.64} & \underline{66.04} & \underline{23.04} & 60.46 & \textbf{46.93} & 64.98 & \underline{75.03} & \textbf{37.11} & \underline{53.37}     \\
\bottomrule
\end{tabular}
\end{adjustbox}
\end{table}

%% file: table/hyperparam.tex
\begin{table}[h!]
\centering
\caption{Effect of learning rate (LR) and weight decay (WD) on the performance of Llama-3.2-1B with MIDUS-HML}
\label{tab:lr_wd}
\begin{adjustbox}{width=0.6\linewidth}
\setlength{\tabcolsep}{4pt}
\begin{tabular}{c c c c c c c c c c c}
\toprule
& & \multicolumn{1}{c}{PPL $\downarrow$} & \multicolumn{8}{c}{Zero-shot Accuracy $\uparrow$} \\
\cmidrule(lr){3-3}\cmidrule(lr){4-11}
LR & WD & Wiki & ARC & LogiQA & Wino & CSQA & BoolQ & PIQA & MMLU & Average \\
\midrule
1e-4 &  1e-6 & 11.64 & 66.16 & \textbf{23.20} & \textbf{61.56} & 46.27 & 65.29 & \textbf{75.08} & 36.91 & \textbf{53.50} \\
1e-4 &  0 & 11.64 & \underline{66.25} & 21.97 & 60.77 & 46.60 & \underline{65.32} & \underline{75.03} & 36.99 & 53.27 \\
1e-4 &  1e-2 & \textbf{11.62} & 65.78 & \underline{23.04} & 59.91 & 45.45 & \textbf{65.35} & 74.81 & \textbf{37.25} & 53.09 \\
1e-5 &  1e-6 & 11.64 & 65.57 & 21.51 & \underline{61.01} & \underline{46.85} & 63.58 & 74.48 & 37.12 & 52.87 \\
5e-5 &  1e-6 & \underline{11.63} & \textbf{66.46} & 22.89 & 60.14 & \textbf{47.58} & 64.86 & 74.54 & \underline{37.23} & \underline{53.39} \\
\bottomrule
\end{tabular}
\end{adjustbox}
\end{table}

%% file: table/ablation.tex
\begin{table}[h!]
\centering
\caption{Ablation study on the structure of HML.}
\label{tab:ablations}
\begin{adjustbox}{width=0.8\linewidth}
\setlength{\tabcolsep}{4pt}
\begin{tabular}{l c c c c c c c c c}
\toprule
& \multicolumn{1}{c}{PPL $\downarrow$} & \multicolumn{8}{c}{Zero-shot Accuracy $\uparrow$} \\
\cmidrule(lr){2-2}\cmidrule(lr){3-10}
Method & Wiki & ARC & LogiQA & Wino & CSQA & BoolQ & PIQA & MMLU & Average \\
\hline
\rowcolor{gray!30}
 MIDUS-HML & \textbf{11.64} & \underline{66.16} & \underline{23.20} & \textbf{61.56} & 46.27 & \textbf{65.29} & 75.08 & 36.91 & \underline{53.50} \\
w/ BatchNorm1d  & 11.74 & \textbf{66.54 }& 22.43 & 60.77 & \textbf{47.34} & 64.80 & 74.92 & 36.94 & 53.39 \\
w/ LayerNorm & \underline{11.68} & 66.08 & 22.43 & \textbf{61.56} & \underline{46.68} & 65.17 & \underline{75.19} & \textbf{37.34} & 53.49 \\
w/ Residual & \textbf{11.64} & 66.04 & 22.89 & \underline{61.48} & 46.36 & 65.17 & \textbf{75.35} & 36.75 & 53.43 \\
w/ Output Projection & \textbf{11.64} & 66.12 & \textbf{23.66} & 61.09 & 46.52 & \underline{65.23} & 74.86 & \underline{37.07} & \textbf{53.51} \\
\bottomrule
\end{tabular}
\end{adjustbox}
\end{table}

%% file: table/ablation_freeze.tex
\begin{table}[h!]
\centering
\caption{MIDUS-HML under different DUS placement policies. Peak GPU Memory and Time are training requirements. \textit{Top-heavy} places expanded blocks toward the top, \textit{Distributed} interleaves, and \textit{Bottom-heavy} places them near the input.}
\label{tab:dus_policy}
\begin{adjustbox}{width=0.8\linewidth}
\setlength{\tabcolsep}{4pt}
\begin{tabular}{c c c c c c c c c c c c}
\toprule
&  \multicolumn{1}{c}{GB $\downarrow$}  & \multicolumn{1}{c}{s/iter $\downarrow$}   & \multicolumn{1}{c}{PPL $\downarrow$} & \multicolumn{8}{c}{Zero-shot Accuracy $\uparrow$} \\
\cmidrule(lr){2-2} \cmidrule(lr){3-3} \cmidrule(lr){4-4}\cmidrule(lr){5-12}
DUS Policy & GPU Memory & Time & Wiki & ARC & LogiQA & Wino & CSQA & BoolQ & PIQA & MMLU & Average \\
\midrule
\textit{Top-heavy} &  \textbf{24.1} & \textbf{3.89} & \underline{11.63} & \underline{66.12} & \underline{22.27} & \underline{61.25} & \textbf{46.44} & 64.28 & 74.92 & 36.23 & \underline{53.07} \\
\textit{Distributed} &  \underline{28.0} & \underline{4.56} & 11.64 & \textbf{66.16} & \textbf{23.20} & \textbf{61.56} & \underline{46.27} & \textbf{65.29} & \underline{75.08} & \textbf{36.91} & \textbf{53.50} \\
\textit{Bottom-heavy} & 28.6 & 4.67 & \textbf{11.60} & \underline{66.12} & 21.81 & 60.30 & 45.05 & \underline{65.05} & \textbf{75.30} & \underline{36.35} & 52.85 \\
\bottomrule
\end{tabular}
\end{adjustbox}
\end{table}

%% file: figure/abl_bs.tex
\begin{figure}[h!]
    \centering
    \includegraphics[width=0.8\linewidth]{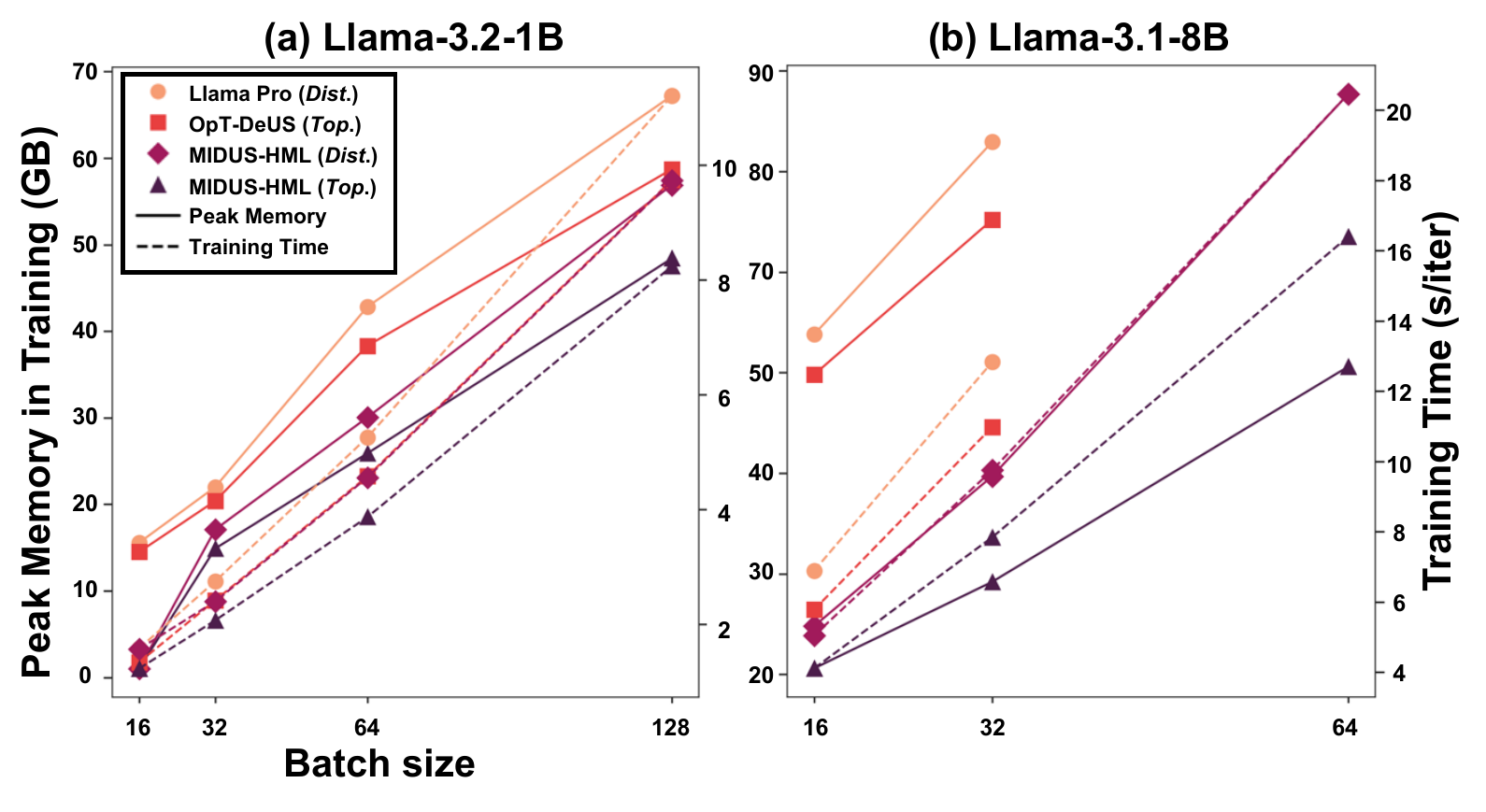}
    \caption{Efficiency of DUS baselines and MIDUS-HML as a function of batch size. \textit{Dist.} denotes the \textit{Distributed} and \textit{Top.} the \textit{Top-heavy} DUS placement policy.}
    \label{fig:abl_bs}
\end{figure}

%% file: figure/abl_seq.tex
\begin{figure}[h!]
    \centering
    \includegraphics[width=0.8\linewidth]{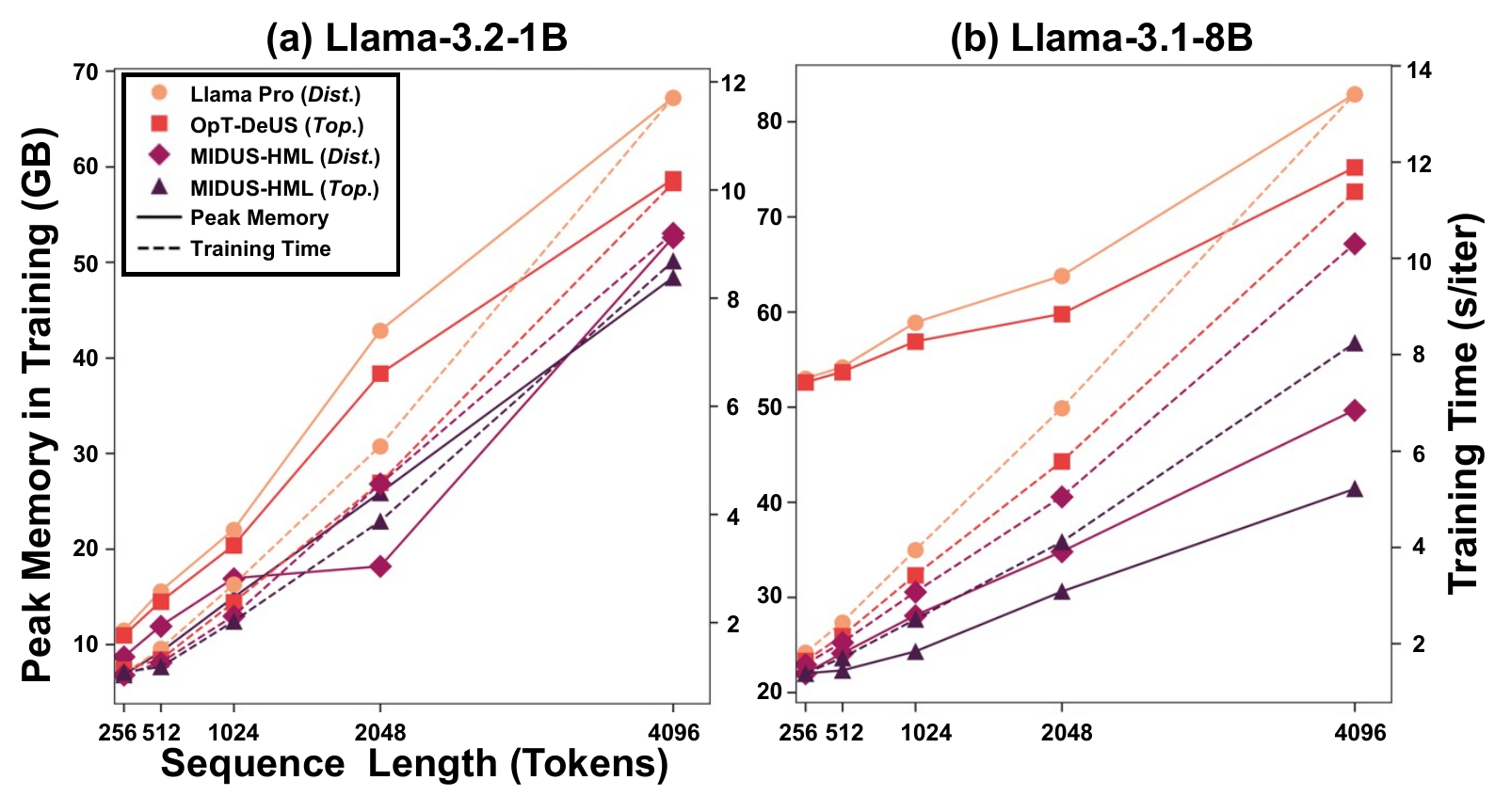}
    \caption{Efficiency of DUS baselines and MIDUS-HML as a function of sequence length. \textit{Dist.} denotes the \textit{Distributed} and \textit{Top.} the \textit{Top-heavy} DUS placement policy.}
    \label{fig:abl_seq}
\end{figure}

%% file: sec/theory_appendix.tex
\subsection{Structural Analysis of Head-wise Value Memory}
\label{app:theory}

This appendix analyzes the head-wise value parameterization used in MIDUS-HML under fixed retrieval coefficients. We focus on the following architectural question. When the desired residual corrections are heterogeneous across heads within a common retrieval span, when does head-specific value capacity reduce the empirical correction loss compared to a shared-realized-value alternative, and when can HIVE preserve this advantage through shared latent storage with head-specific value realization?

\subsubsection{Local Correction Setup}
\label{app:theory_setup}

Consider fixed attention-head outputs
\[
a'=[a_1\mid\cdots\mid a_H]\in\mathbb{R}^{T\times d},
\qquad
a_h\in\mathbb{R}^{T\times d_h},
\]
where $H$ is the number of attention heads, $d=Hd_h$, and $[H]=\{1,\ldots,H\}$. Unless specified otherwise, $\|\cdot\|$ denotes the Euclidean norm for vectors and the Frobenius norm for matrices.

We analyze value updates under fixed retrieval coefficients. For each head $h$, we collect the sparse Top-$k$ retrieval-weight vectors $\alpha_{h,t}\in\Delta^{N-1}$ into the retrieval-weight matrix
\[
A_h=[\alpha_{h,1},\dots,\alpha_{h,T}]^\top\in\mathbb{R}^{T\times N},
\]
where $N$ is the number of composite memory slots per head. Let $Y_h\in\mathbb{R}^{T\times d_h}$ denote the desired residual correction for head $h$, and let
\[
Y=[Y_1\mid\cdots\mid Y_H]\in\mathbb{R}^{T\times d}.
\]
For a head-partitioned correction
\[
F(a')=[F_1(a')\mid\cdots\mid F_H(a')],
\]
define the empirical correction loss
\[
\mathcal{L}(F)
=
\frac12\sum_{h=1}^{H}\|F_h(a')-Y_h\|^2.
\]

We compare two value parameterizations under the same retrieval-weight matrices $\{A_h\}_{h=1}^{H}$.

\paragraph{Shared realized value bank.}
The shared-realized-value family uses a single realized value bank $V_{\text{sh}}\in\mathbb{R}^{N\times d_h}$ for all heads and is defined by
\[
F_h(a')=A_hV_{\text{sh}}.
\]

\paragraph{Head-specific value banks.}
The head-specific-value family assigns an independent value bank $V_h\in\mathbb{R}^{N\times d_h}$ to each head and is defined by
\[
F_h(a')=A_hV_h.
\]

\subsubsection{Loss Gap from Head-specific Value Capacity}
\label{app:loss_gap}
For fixed retrieval, both families are linear in their value parameters. The optimal head-specific-value loss is
\[
\mathcal{L}_{\mathrm{Ind}}^\star
=
\inf_{\{V_h\}_{h=1}^{H}}\mathcal{L}(F)
=
\frac12\sum_{h=1}^{H}\|(I_T-\Pi_{A_h})Y_h\|^2,
\]
where $\Pi_{A_h}$ is the orthogonal projector onto $\operatorname{col}(A_h)\subseteq\mathbb{R}^{T}$. In contrast, the shared-realized-value family must use one common bank $V_{\text{sh}}$ to fit all heads simultaneously.

\begin{proposition}\label{prop:shared_vs_specific}
The shared-realized-value optimum is
\begin{equation}
\mathcal{L}_{\mathrm{Share}}^\star
=
\inf_{V_{\text{sh}}}\mathcal{L}(F)
=
\mathcal{L}_{\mathrm{Ind}}^\star
+
\frac12
\sum_{h=1}^{H}
\|A_hV_{\text{sh}}^\star-\Pi_{A_h}Y_h\|^2,
\label{eq:proposition1}
\end{equation}
where
\[
V_{\text{sh}}^\star
=
\Bigl(\sum_{h=1}^{H}A_h^\top A_h\Bigr)^{\dagger}
\sum_{h=1}^{H}A_h^\top Y_h
\]
is a minimum-norm optimal shared realized value bank, and $\dagger$ denotes the Moore--Penrose pseudoinverse. Define $\Delta_{\mathrm{Share}}:=\mathcal{L}_{\mathrm{Share}}^\star-\mathcal{L}_{\mathrm{Ind}}^\star$. Then $\Delta_{\mathrm{Share}}\ge 0$. Moreover, $\Delta_{\mathrm{Share}}=0$ if and only if there exists a single $V_{\text{sh}}\in\mathbb{R}^{N\times d_h}$ such that $A_hV_{\text{sh}}=\Pi_{A_h}Y_h$ for all $h\in[H]$. When retrieval is identical across heads, that is, $A_h=A$ for all $h$, the gap reduces to
\[
\Delta_{\mathrm{Share}}
=
\frac12
\sum_{h=1}^{H}
\|\Pi_A(Y_h-\bar Y)\|^2,
\qquad
\bar Y=\frac1H\sum_{h=1}^{H}Y_h.
\]
Thus, in this special case, the shared-realized-value penalty is the sum of squared deviations of head-specific correction targets within the common retrieval span $\operatorname{col}(A)$.
\end{proposition}

\begin{proof}
The head-specific optimum follows by solving, independently for each head,
\[
\inf_{V_h}\frac12\|A_hV_h-Y_h\|^2,
\]
which gives the projection residual
\[
\frac12\|(I_T-\Pi_{A_h})Y_h\|^2.
\]

For the shared-realized-value family, the optimization is
\[
\inf_{V_{\text{sh}}}
\frac12
\sum_{h=1}^{H}
\|A_hV_{\text{sh}}-Y_h\|^2.
\]
The normal equation is
\[
\Bigl(\sum_{h=1}^{H}A_h^\top A_h\Bigr)V_{\text{sh}}
=
\sum_{h=1}^{H}A_h^\top Y_h,
\]
and the minimum-norm solution is
\[
V_{\text{sh}}^\star
=
\Bigl(\sum_{h=1}^{H}A_h^\top A_h\Bigr)^{\dagger}
\sum_{h=1}^{H}A_h^\top Y_h.
\]

For each head, $A_hV_{\text{sh}}^\star$ and $\Pi_{A_h}Y_h$ both have columns in $\operatorname{col}(A_h)$, whereas $(I_T-\Pi_{A_h})Y_h$ has columns in $\operatorname{col}(A_h)^\perp$. Hence $A_hV_{\text{sh}}^\star-\Pi_{A_h}Y_h$ is orthogonal to $(I_T-\Pi_{A_h})Y_h$ column-wise. Since
\[
A_hV_{\text{sh}}^\star-Y_h
=
(A_hV_{\text{sh}}^\star-\Pi_{A_h}Y_h)
-
(I_T-\Pi_{A_h})Y_h,
\]
the Pythagorean identity yields
\[
\|A_hV_{\text{sh}}^\star-Y_h\|^2
=
\|A_hV_{\text{sh}}^\star-\Pi_{A_h}Y_h\|^2
+
\|(I_T-\Pi_{A_h})Y_h\|^2.
\]
Applying this decomposition to the shared objective gives Eq.~\eqref{eq:proposition1}. The non-negativity and equality condition follow immediately.

If $A_h=A$ for all $h$, then
\[
V_{\text{sh}}^\star=(A^\top A)^\dagger A^\top \bar Y,
\qquad
AV_{\text{sh}}^\star=\Pi_A\bar Y.
\]
Substituting this into the gap expression gives
\[
\frac12
\sum_{h=1}^{H}
\|\Pi_A\bar Y-\Pi_AY_h\|^2
=
\frac12
\sum_{h=1}^{H}
\|\Pi_A(Y_h-\bar Y)\|^2.
\]
\end{proof}

Proposition~\ref{prop:shared_vs_specific} quantifies the penalty of forcing all heads to use a single realized value bank. The gap vanishes only when a common bank can match every projected head-specific target.

\subsubsection{HIVE as Shared Latent Storage with Head-specific Value Projection}
\label{app:hive_approximation}

We now consider HIVE, which shares a latent value table $\bar V\in\mathbb{R}^{N\times r}$ while allowing each head to use its own value projection $W_h\in\mathbb{R}^{r\times d_h}$, so that the head-specific bank is represented as $\bar V W_h$.

The next proposition shows that if an optimal head-specific solution is approximately generated from a shared latent value bank with head-specific value projections, then HIVE inherits the independent head-specific loss up to a controlled residual penalty.

\begin{proposition} \label{prop:hive_latent_compressibility}
Let $\{V_h^\star\}_{h=1}^{H}\in\arg\min_{\{V_h\}_{h=1}^{H}}\mathcal{L}(F)$ be a selected optimal head-specific solution. Suppose there exist $\bar V\in\mathbb{R}^{N\times r}$, $W_h\in\mathbb{R}^{r\times d_h}$, and $R_h\in\mathbb{R}^{N\times d_h}$ such that
\[
V_h^\star=\bar V W_h+R_h
\qquad
\text{for all }h\in[H],
\]
and $\sum_{h=1}^{H}\|R_h\|^2\le \varepsilon$, where $r\le d_h$. Let $\rho=\max_{h\in[H]}\|A_h\|_{\text{op}}^2$, where $\|\cdot\|_{\text{op}}$ denotes the operator norm.  Then the corresponding HIVE correction $\widehat F_h(a')=A_h\bar V W_h$ satisfies
\[
\mathcal{L}(\widehat F)
\le
\mathcal{L}_{\mathrm{Ind}}^\star+\frac{\rho}{2}\varepsilon.
\]
\end{proposition}

\begin{proof}
Define
\[
E_h:=V_h^\star-\bar V W_h.
\]
By assumption, $E_h=R_h$. Let
\[
F_h^\star(a'):=A_hV_h^\star.
\]
The corresponding HIVE correction satisfies
\[
\widehat F_h(a')-F_h^\star(a')=-A_hE_h.
\]
By the operator-Frobenius bound,
$$\|\widehat F_h(a')-F_h^\star(a')\|^2 = \|A_hE_h\|^2 \le \|A_h\|_{\text{op}}^2 \|E_h\|^2 \le \rho\|E_h\|^2.$$

Since $V_h^\star$ is optimal for the head-specific least-squares problem, its fitted value is the column-wise projection
\[
F_h^\star(a')=A_hV_h^\star=\Pi_{A_h}Y_h.
\]
Hence $F_h^\star(a')-Y_h$ has columns in $\operatorname{col}(A_h)^\perp$, while $A_hE_h$ has columns in $\operatorname{col}(A_h)$. These two terms are orthogonal, and therefore
\[
\|\widehat F_h(a')-Y_h\|^2
=
\|F_h^\star(a')-Y_h\|^2
+
\|\widehat F_h(a')-F_h^\star(a')\|^2.
\]
Summing over heads gives
\[
\mathcal{L}(\widehat F)
\le
\mathcal{L}_{\mathrm{Ind}}^\star
+
\frac{\rho}{2}\sum_{h=1}^{H}\|E_h\|^2.
\]
Since $E_h=R_h$ and $\sum_{h=1}^{H}\|R_h\|^2\le\varepsilon$, we obtain
\[
\mathcal{L}(\widehat F)
\le
\mathcal{L}_{\mathrm{Ind}}^\star+\frac{\rho}{2}\varepsilon.
\]
\end{proof}

We now prove Theorem~\ref{thm:theory1}.

\structuredheterogeneity*

\begin{proof}[Proof]
By the assumed shared latent factorization,
\[
V_h^\star=\bar V W_h
\qquad
\text{for all } h\in[H].
\]
This is a special case of Proposition~\ref{prop:hive_latent_compressibility} with
\[
R_h=0
\qquad
\text{for all }h\in[H],
\]
and hence $\varepsilon=0$. Therefore,
\[
\mathcal{L}(\widehat F)
\le
\mathcal{L}_{\mathrm{Ind}}^\star.
\]
Since every HIVE correction corresponds to a valid head-specific value choice with value banks $\bar V W_h$, it also belongs to the head-specific-value family. Therefore,
\[
\mathcal{L}(\widehat F)
\ge
\mathcal{L}_{\mathrm{Ind}}^\star.
\]
Thus,
\[
\mathcal{L}(\widehat F)=\mathcal{L}_{\mathrm{Ind}}^\star.
\]

It remains to compare this value with the shared-realized-value optimum. Under the identical-retrieval assumption $A_h=A$ for all $h$, Proposition~\ref{prop:shared_vs_specific} gives
\[
\mathcal{L}_{\mathrm{Share}}^\star
=
\mathcal{L}_{\mathrm{Ind}}^\star
+
\frac12
\sum_{h=1}^{H}
\|\Pi_A(Y_h-\bar Y)\|^2,
\qquad
\bar Y=\frac1H\sum_{h=1}^{H}Y_h.
\]
The assumption
\[
\Pi_AY_h\neq\Pi_AY_{h'}
\qquad
\text{for some }h\neq h'
\]
implies that not all projected targets are equal to their mean $\Pi_A\bar Y$. Hence
\[
\sum_{h=1}^{H}
\|\Pi_A(Y_h-\bar Y)\|^2
>
0.
\]
Therefore,
\[
\mathcal{L}_{\mathrm{Share}}^\star
>
\mathcal{L}_{\mathrm{Ind}}^\star.
\]
Combining this with $\mathcal{L}(\widehat F)=\mathcal{L}_{\mathrm{Ind}}^\star$ yields
\[
\mathcal{L}(\widehat F)
=
\mathcal{L}_{\mathrm{Ind}}^\star
<
\mathcal{L}_{\mathrm{Share}}^\star.
\]
\end{proof}

The theorem isolates a structured heterogeneity regime in which HIVE matches the head-specific optimum. The condition $\Pi_AY_h\neq\Pi_AY_{h'}$ for some $h\neq h'$ means that different heads require different projected corrections even under a common retrieval span, making a single shared realized value bank restrictive. Meanwhile, the shared latent factorization $V_h^\star=\bar V W_h$ is the zero-residual instance of Proposition~\ref{prop:hive_latent_compressibility}, where the optimal head-specific banks share a common latent value table. HIVE matches this pattern by sharing latent storage while allowing head-specific value realization, thereby separating storage-level sharing from head-specific correction.

\paragraph{Connection to FFN-style correction blocks.}
Prior work interprets Transformer FFNs as implicit key--value memories, where the first linear map and nonlinearity induce key-like activations and the second linear map provides value-like outputs~\citep{geva2021transformer}. Under this view, duplicating FFN branches in DUS can be interpreted as increasing memory-like value capacity through a shared block-level correction over the full hidden representation.

The analysis above should not be read as a formal reduction of arbitrary FFNs to the shared-realized-value family, since FFNs include nonlinear activations and dense feature mixing. Instead, it isolates a simplified fixed-retrieval regime that reflects the distinction between shared realized values and head-specific value realization. HML with HIVE shares a latent value table while allowing each head to realize this shared storage through its own value projection. The theorem shows that this separation can matter when heads share the same retrieval-weight matrix but require different projected corrections.